\documentclass[10pt,twocolumn,letterpaper]{article}

\usepackage{iccv}              % To produce the CAMERA-READY version
\usepackage{soul}
\usepackage[utf8]{inputenc} % allow utf-8 input
\usepackage[T1]{fontenc}    % use 8-bit T1 fonts
\usepackage{url}            % simple URL typesetting
\usepackage{booktabs}       % professional-quality tables
\usepackage{amsfonts}       % blackboard math symbols
\usepackage{nicefrac}       % compact symbols for 1/2, etc.
\usepackage{microtype}      % microtypography
\usepackage{graphicx}
\usepackage{epstopdf}
\usepackage{multirow}
\usepackage{xstring}
\usepackage{arydshln}
\usepackage{float}
\usepackage{amsmath}
\usepackage{bm}
\usepackage[accsupp]{axessibility}
\definecolor{iccvblue}{rgb}{0.21,0.49,0.74}
\usepackage[pagebackref,breaklinks,colorlinks,allcolors=iccvblue]{hyperref}

\def\paperID{5846} % *** Enter the Paper ID here
\def\confName{ICCV}
\def\confYear{2025}

\title{Visual Textualization for Image Prompted Object Detection}

\author{
Yongjian Wu$^1$\thanks{Authors contributed equally.}\;\;\;\; 
Yang Zhou$^1$\footnotemark[1]\;\;\;\; 
Jiya Saiyin$^1$\;\;\;\; 
Bingzheng Wei$^2$\;\;\;\;
Yan Xu $^1$\thanks{Corresponding author.}\\
$^1$School of Biological Science and Medical Engineering, Beihang University\;\;\; \\$^2$ByteDance Inc.\\
{\tt\small {wuyongjian,zhouyangbme}@buaa.edu.cn}\;\;\;  {\tt\small \{xuyan04\}@gmail.com}
}

\begin{document}
\maketitle

\begin{abstract}
%Object-level vision-language models (OVLMs) exhibit strong object feature representation capabilities and highly generalized object-text alignment, making them effective for zero-shot object detection and well-suited for building advanced few-shot object detection (FSOD) systems. However, conventional methods that incorporate support image information by fine-tuning weights or adding additional processing structures often disrupt OVLM's well-established object-text alignment, adversely affecting FSOD generalization performance.
\textcolor{black}{We propose VisTex-OVLM, a novel image prompted object detection method that introduces visual textualization —-- a process that projects a few visual exemplars into the text feature space to enhance Object-level Vision-Language Models' (OVLMs) capability in detecting rare categories that are difficult to describe textually and nearly absent from their pre-training data, while preserving their pre-trained object-text alignment. Specifically, VisTex-OVLM leverages multi-scale textualizing blocks and a multi-stage fusion strategy to integrate visual information from visual exemplars, generating textualized visual tokens that effectively guide OVLMs alongside text prompts. Unlike previous methods, our method maintains the original architecture of OVLM, maintaining its generalization capabilities while enhancing performance in few-shot settings. VisTex-OVLM demonstrates superior performance across open-set datasets which have minimal overlap with OVLM's pre-training data and achieves state-of-the-art results on few-shot benchmarks PASCAL VOC and MSCOCO}. The code will be released at \href{https://github.com/WitGotFlg/VisTex-OVLM}{VisTex-OVLM}.
\end{abstract}

% , aiming at detecting novel objects with only a limited number of annotated images --- referred to as support images in the FSOD setting --- striving to approximate human-level recognition ability

\section{Introduction}
%\textbf{However, as for humans, even a child can recognize novel classes after only a few exposures.} , a fundamental task in computer vision (CV) involving spatial localization and semantic recognition,
% Consequently, researchers have turned to few-shot object detection (FSOD) \cite{Bulat_2023fsdetr, Han_2024_FM-FSOD, wang2019metadet, Qiao_2021_defrcn2, han2022fct}.
% Current best-performed deep learning object detection methods \cite{redmon2016you,ren2016faster} remain dependent on large labeled datasets, which are often unavailable in contexts like rare species detection or medical applications. Consequently, researchers have turned to develop object detection methods that only require a few visual exemplars, which also refer to few-shot object detection (FSOD) \cite{Bulat_2023fsdetr, Han_2024_FM-FSOD, wang2019metadet, Qiao_2021_defrcn2, han2022fct}. 
Recently, vision-language models (VLMs) have demonstrated exceptional generalization in CV by pre-training on extensive image-text pairs, allowing zero-shot transfer via natural language prompts and inspiring new approaches for object detection \cite{guo2024DP-DDCL, han2022mmfsod, xu2023normvae, zhao2024veic, li2023dr, Han_2024_FM-FSOD}. Among VLMs, object-level VLMs (OVLMs) \cite{li2022glip, dou2022fiber, liu2023groundingdino}, represented by GLIP \cite{li2022glip}, achieve better object-text alignment. \textcolor{black}{OVLMs are pre-trained with extensive object detection and phrase grounding data, featuring multi-stage and multi-scale encoders to enhance object-text alignment beyond general VLMs.} Furthermore, OVLMs commonly employ a multi-stage encoding structure based on cross-attention, allowing text prompts to actively guide object feature representation and location regression. These characteristics enable OVLMs to surpass conventional VLMs like CLIP \cite{radford2021clip} in zero-shot transfer for downstream object detection tasks. % GLIP \cite{li2022glip} is a representative work of OVLMs.

However, OVLM's zero-shot object detection (ZSOD) has intrinsic limitations. First, many objects in downstream tasks are underrepresented in the pre-training data, leading to suboptimal zero-shot transfer performance. Second, text prompts often lack sufficient description granularity, resulting in semantic bias and omissions. Third, fine-grained objects share similar descriptions in the text space, making it difficult to distinguish them using text alone \cite{xu2023mqdet,du2022learning}. These challenges make introducing new visual information from downstream tasks essential for improving OVLM's transfer ability. One solution is to leverage visual exemplars for fine-tuning, which also refers to few-shot object detection (FSOD) \cite{Bulat_2023fsdetr, Han_2024_FM-FSOD, wang2019metadet, Qiao_2021_defrcn2, han2022fct}. For example, some VLM-based object detection methods incorporate few-shot visual exemplars information by fine-tuning original weights or adding processing structures. However, these approaches risk disrupting the OVLM's original object-text alignment and harm its generalization capabilities.

\begin{figure}[t]
  \centering
  \includegraphics[width=0.96\columnwidth]{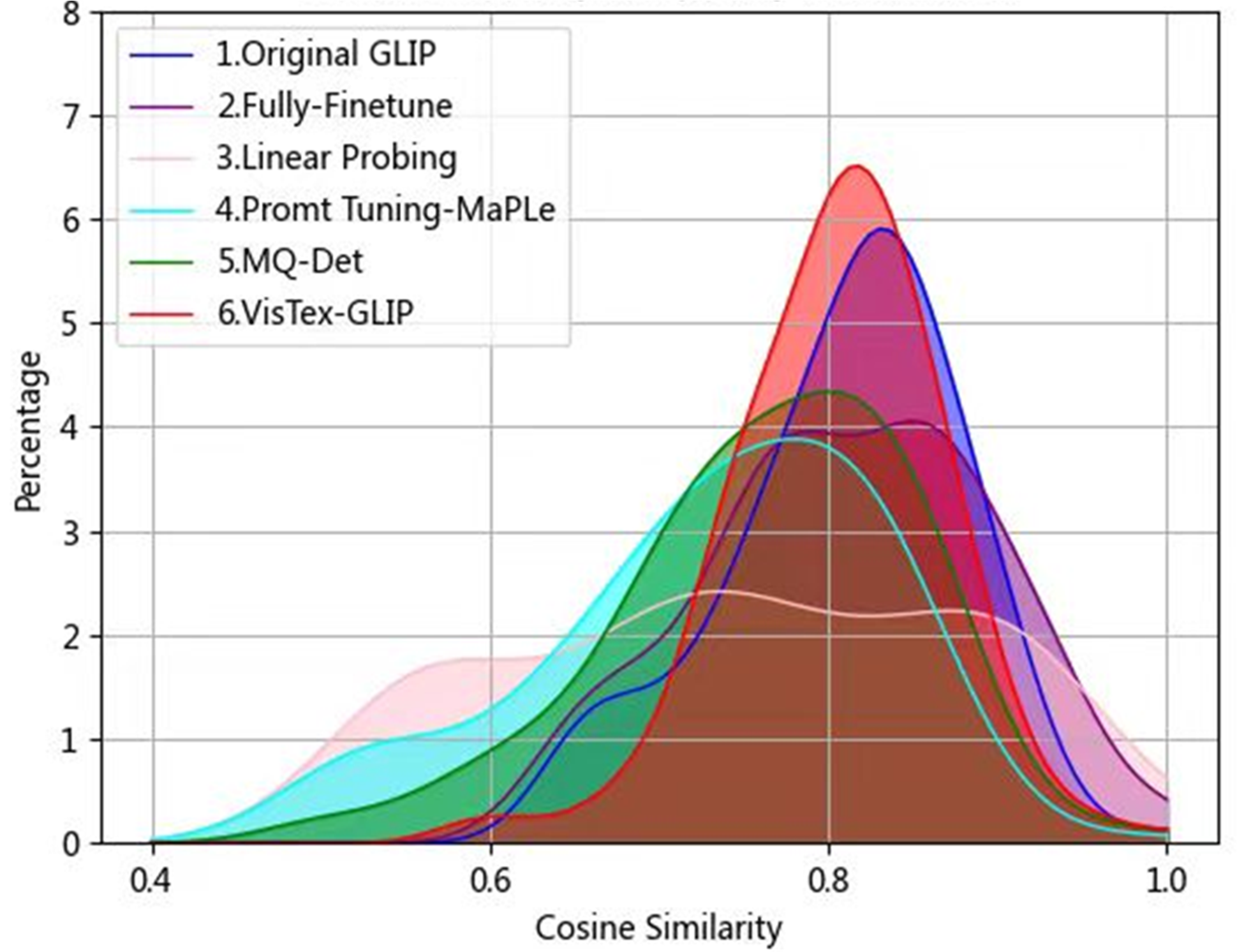}
  \caption{\textcolor{black}{Frequency distribution of feature cosine similarity between object-text pairs from Object365 dataset after transferring the Object365-pretrained GLIP to MSCOCO with different methods.}}
  \vskip -0.08in
  \label{fig:cosine}
\end{figure}

To illustrate this, without loss of generality, we conducted an empirical analysis by transferring GLIP, pre-trained on Object365 \cite{shao2019objects365}, to MSCOCO \cite{lin2014microsoftCOCO} using various common transfer methods. \textcolor{black}{Post-transfer, we computed the cosine similarity between paired text and object features from Object365} (\cref{fig:cosine}). \textcolor{black}{Results revealed that any modification in model structure or weights (methods 2-4) with limited transfer data skewed the object-text similarity distribution in the source domain, partially degrading alignment.} This raises the question: can we introduce the semantics of a few visual exemplars while preventing the limited target training data from disrupting OVLM’s object-text alignment?

Using images as prompts, i.e., image prompting \cite{xu2023mqdet,luddecke2022clipseg}, to provide semantic guidance alongside text prompts may address the question above. \textcolor{black}{However, most OVLMs \cite{li2022glip, dou2022fiber, liu2023groundingdino} are text-prompt-exclusive, lacking native support for image prompting due to architectural incompatibilities.} MQ-Det \cite{xu2023mqdet} is the first to integrate visual exemplars into OVLM's prompting process by adding extra trainable cross-attention modules within the text encoder, enabling exemplar images to modulate the text prompts token-wisely and introduce the visual information of novel classes during inference. However, MQ-Det only re-weights tokens of existing text prompts without directly introducing new visual information, which may result in the loss of key clues. Moreover, the newly added cross-attention structures also bias OVLM's pre-trained object-text alignment, as shown in method 5, \cref{fig:cosine}. \textcolor{black}{How to enable image prompting for text-prompt-exclusive OVLMs without disrupting their pre-trained object-text alignment remains underexplored.}

This paper proposes VisTex-OVLM, a novel image prompting method for expanding OVLM's detection capacity beyond pre-trained categories. To preserve OVLM’s object-text alignment, we introduce visual textualization, which projects support objects into the text feature space. This process allows limited exemplars from novel categories to participate in prompting, guiding OVLM to perform open-set detection alongside semantic text prompts. The visual textualized support images, combined with the text prompt, are directly input into the unmodified pre-trained OVLM, maintaining its established object-text alignment. Specifically, we first leverage OVLM’s visual encoder to extract representations from visual exemplars and design lightweight multi-scale textualizing blocks (MSTBs) for projection. MSTBs process the multi-scale intermediate visual features of the given exemplars produced by each layer of OVLM’s visual encoder and uniformly project them into OVLM’s text feature space as textualized visual features. Furthermore, we employ a non-parametric multi-stage fusion strategy (MSF) to fuse the textualized visual features from different stages of the visual encoder, \textcolor{black}{leveraging OVLM’s multi-stage object-text alignment without introducing additional parameters}. Each visual exemplar is projected into a textualized visual token, guiding OVLM in inferring the target image alongside the text prompts. \textcolor{black}{MSTB and MSF fully leverage an OVLM’s original structure, robust object-text alignment, and its capability for object-level feature extraction, enabling the textualized visual tokens to achieve optimal semantic representation.} During training, MSTB is the only trainable structure. Without fine-tuning on novel classes, MSTB can directly project the visual exemplars from novel categories into semantic-rich textualized visual tokens, \textcolor{black}{which is suitable for practical applications such as open-vocabulary detection on personal mobile devices using image prompts.} \textcolor{black}{Previous methods break OVLMs’ pre-trained object-text alignment when introducing image prompts. VisTex-OVLM's novelty lies in effectively leveraging the original OVLM structure to incorporate visual information that complements underspecified semantics missing from the text prompts, without altering the model’s pre-trained alignment.}

Unlike MQ-Det which modulates text features, VisTex-OVLM directly incorporates visual information, preserving OVLM’s inference structure and object-text alignment \textcolor{black}{to the greatest extent (method 6, \cref{fig:cosine})}. \textcolor{black}{When applied to GLIP (VisTex-GLIP) and GroundingDINO \cite{liu2023groundingdino} (VisTex-DINO), VisTex-OVLM demonstrates superior performance in open-set scenarios, including LVIS \cite{gupta2019lvis} and 16 datasets with minimal overlap with OVLM pre-training data. Furthermore, VisTex-OVLM also achieves state-of-the-art (SOTA) results on standard few-shot benchmarks, including PASCAL VOC \cite{everingham2010pascalvoc} and MSCOCO \cite{lin2014microsoftCOCO}, showing broad applicability.} Extensive ablation studies further validate the effectiveness of using images as semantic complements to text prompts. The contributions are summarized as follows:
\begin{itemize}
    \item We propose VisTex-OVLM, introducing visual textualization, which projects visual exemplars into the text feature space for image prompting to expand OVLM’s practical applicability in detecting categories absent from pre-training data without disrupting their object-text alignment.
    \item Multi-scale textualizing blocks (MSTBs) and a multi-stage fusion (MSF) strategy are designed to preserve the original structure and robust alignment of OVLMs while enabling efficient utilization of their pre-trained knowledge.
    %We design multi-scale textualizing blocks (MSTBs) and a multi-stage fusion (MSF) strategy for efficient visual textualization, leveraging OVLM's object-level feature extraction to achieve optimal semantic representation for textualized visual tokens.
    \item VisTex-OVLM excels with limited visual exemplars in open-set scenarios and achieves SOTA performance on standard FSOD benchmarks. % by adding image prompts
\end{itemize}

\section{Related works}
\subsection{Few-shot object detection}

\begin{figure*}[t]
  \centering
  \includegraphics[width=0.96\textwidth]{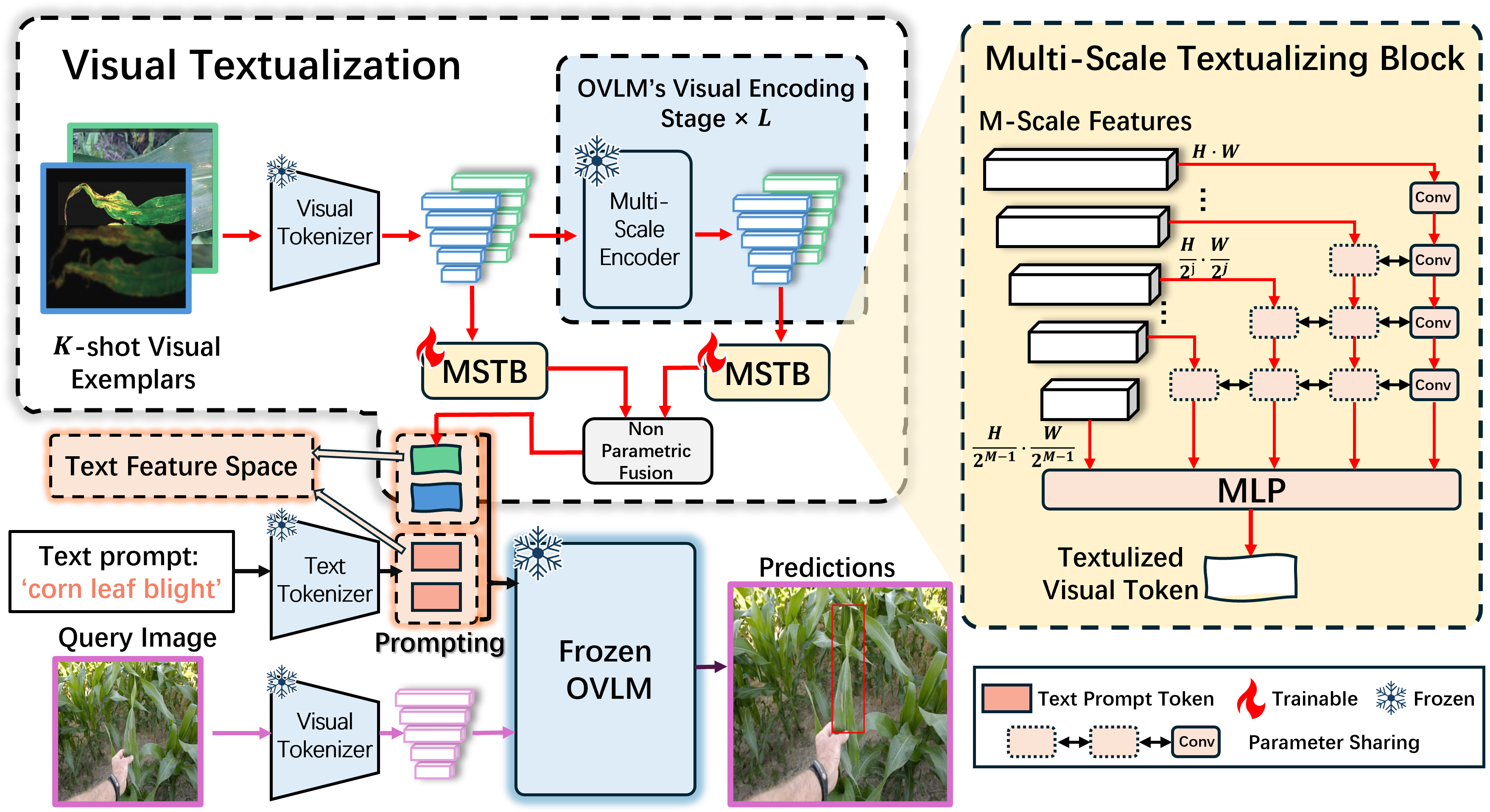}
  \caption{Overview of VisTex-OVLM. VisTex-OVLM performs visual textualization of support images through parameter-shared multi-scale textualizing blocks (MSTB) and a non-parametric multi-stage fusion strategy, mapping support images to text feature space for direct prompting of unmodified OVLMs, such as GLIP \cite{li2022glip} and GroundingDINO \cite{liu2023groundingdino}.}
  % \vskip -0.14in
  \label{fig:fig2}
\end{figure*}

Few-shot object detection aims at detecting novel objects utilizing only few-shot annotated instances from novel classes. Mainstream methods mostly rely on complex episode-based meta-learning training paradigms \cite{zhang2022metaDETR1, han2022meta-faster-rcnn1, fan2020-1, han2021query-1, han2022-1} or intricate inter-class relationship modeling \cite{Qiao_2021_defrcn2, wang2020tfa2, wu2022mfd2, wang2024snida2}. Despite the progress made, FSOD remains a challenging task. Recently, VLMs have shown remarkable generalization capabilities, which led to their growing application in FSOD \cite{guo2024DP-DDCL, han2022mmfsod, xu2023normvae, zhao2024veic, li2023dr, Han_2024_FM-FSOD}. However, most approaches demonstrate limited performance. The current SOTA model, MTL-FSOD \cite{ren2024MTL-FSOD}, leverages VLMs solely for enhancing classification in FSOD through knowledge distillation, falling short of fully exploiting the potential of VLMs in FSOD. Our VisTex-OVLM probes deeper, aiming to fully leverage the potential of VLM in FSOD. 

\subsection{VLM-based detection and grounding}

VLMs leverage vast text-image pairs obtained from the internet to train its text encoder and image encoder with contrastive learning, allowing the model to learn highly generalizable features. Among them, CLIP \cite{radford2021clip} is widely recognized for its image-level alignment, supporting effective transfer to image classification and text-image retrieval tasks. However, due to its lack of object-level knowledge, CLIP requires fine-tuning or adding detector modules for detection and grounding tasks \cite{kuo2022fvlm, gu2021vild, wu2023cora, zhao2024samp}, increasing model complexity and training costs. Object-level VLMs (OVLMs) designed for detection tasks have emerged to address this \cite{li2022glip, dou2022fiber, liu2023groundingdino}. \textcolor{black}{A representative work, GLIP \cite{li2022glip} combines object detection and phrase grounding data, utilizing grounded pre-training and a multi-scale, multi-stage cross-attention architecture to establish strong object-text alignment, leading to superior downstream detection and grounding performance over image-level open-vocabulary detectors like regionCLIP \cite{zhong2022regionclip}.} Therefore, our VisTex-OVLM is built upon OVLMs, which are inherently well-suited for detection and grounding tasks, to further explore the potential of VLM on FSOD. 

\subsection{Prompting methods}

Prompting provides task-specific hints to guide large pre-trained models, enhancing their understanding and results \cite{gu2023systematic}. Originating in NLP, it now extends to VLMs.

\textbf{Design and tuning} Prompt design refers to manually constructing prompts \cite{efrat2020turking, radford2019language}, which can be complex, time-consuming, and prone to subjective biases. Prompt tuning adds learnable tokens to VLMs and fine-tunes \cite{bahng2022vp, chowdhury2023apollo, jia2022vpt, khattak2023pt, zhou2022cocoop, zhou2022coop}. However, tuning may disrupt VLMs' cross-modal alignment, defecting its generalization capabilities.

\textbf{Image prompting} Image prompting is a novel method where images are regarded as prompts to provide semantic guidance alongside text prompts. CLIPSeg \cite{luddecke2022clipseg} applies this for one-shot segmentation but requires an extra segmentation module, which can disrupt cross-modal alignment. In detection tasks, \textcolor{black}{OWL-ViT \cite{minderer2022simple} allows image prompting, but it relies on image-object pre-training and a pre-defined architecture and thus cannot be generalized to other OVLMs.} MQ-Det \cite{xu2023mqdet} integrates a trainable cross-attention module within the text encoder of OVLM, allowing the image prompt to modulate the text prompt in a token-wise manner and thereby introducing the information of prompt images. However, MQ-Det only re-weights tokens without fully incorporating new visual information. Moreover, the additional cross-attention structure may also compromise OVLM’s object-text alignment and generalization.

In contrast to the aforementioned approaches, our VisTex-OVLM leverages few-shot support images as prompts through visual textualization, inserting their semantics directly into OVLM’s text features. This allows the visual information of image prompts to be directly incorporated into the text prompt. More importantly, our approach does not alter the original structure of OVLM, fully maintaining its excellent object-text alignment, which results in excellent generalization ability and performance. 

\section{Method}
\textcolor{black}{Without loss of generality, we introduce our approach using notations of few-shot object detection (FSOD).}

\subsection{Task definition}
In the classical few-shot object detection (FSOD) setting, there are two distinct datasets: a base set $D_b$ with category set $C_b$ and a novel set $D_n$ with category set $C_n$, where $C_b\cap C_n=\emptyset$. The base set $D_b$ contains abundant annotated objects that can be utilized for training the FSOD system, while in the novel set $D_n$, each category $C\in C_n$ has at most $K$ bounding box annotations for $K$-shot (e.g., $K = 1, 2, 3...10$) object detection learning. In the inference phase, the images providing $K$-shot annotations are referred to as support images,while the images to be predicted, termed query images, belong to the category set $C_n$. 

In the more practical generalized few-shot object detection (GFSOD) setting, the target category set during inference becomes $C_b\cup C_n$, requiring the model to maintain robust performance on base classes as well.

\subsection{Preliminaries and notations for OVLMs}

Commonly, OVLMs extend conventional VLMs by incorporating specialized designs for object-level visual tasks \cite{li2022glip, dou2022fiber, liu2023groundingdino}. \textcolor{black}{First, they use a hierarchical visual encoder to perform multi-scale encoding and generate region features at different spatial scales—aligned with text features—to capture objects of varying sizes. Second, they expand the text encoder’s vocabulary using web-sourced phrase grounding data. Third, they adopt a multi-stage architecture with cross-modal attention to enhance region-text alignment.} These specialized designs enable OVLMs to achieve superior object-text alignment, making them particularly suitable for object detection tasks. % Therefore, this paper builds the FSOD system upon GLIP.

To facilitate subsequent discussions, we now define the notation for the OVLM encoding process. Let the intermediate visual features and text features in OVLMs be denoted as $R^i$ and $P^i$, respectively, where $i =0, 1, …, L$, and $L$ is the total number of stages. $R^0$ denotes the visual tokens outputted from a pre-trained visual tokenizer, typically a Swin \cite{liu2021swin}, $P^0$ denotes the text prompt tokens tokenized by a pre-trained text tokenizer, usually BERT \cite{kenton2019bert}. Suppose the text prompt $t$ for target objects contains $\lvert C\rvert$ classes, and for simplicity, assume each class is described using $N$ words ($N=1$ if only the class name is used as the prompt). Throughout the encoding process, $P\in \mathbb{R}^{\lvert C\rvert \cdot N\times d_{T}}$, where $d_T$ is the dimension of the text feature space. Due to the multi-scale encoding, $R^i = \left[ R^{i,(0)}, ..., R^{i,(j)}, ..., R^{i,(M-1)}\right]=\left[R^{i,(j)}\right]_{j=0}^{M-1}$, where $R^{i,(j)}\in \mathbb{R}^{\frac{H}{2^j}\cdot \frac{W}{2^j} \times d_I }$, $H$ and $W$ are the height and width of the largest scale visual features, $M$ is the total number of scales, and $d_I$ is the visual feature space dimension. The notation $\left[\cdot\right]$ represents concatenation.

\subsection{VisTex-OVLM}

% As mentioned in the introduction, using solely text to handle complex downstream tasks in real-world scenarios presents significant challenges for OVLMs, particularly in terms of domain gap and insufficient description granularity.
As revealed by the empirical analysis in the introduction, the fine-tuning of weights or additional processing structures tends to compromise an OVLM’s well-established object-text alignment, adversely affecting the generalization. To address these challenges, we propose VisTex-OVLM, a novel approach that textualizes support images as prompts during the inference phase to guide a non-finetuned pre-trained OVLM in detecting novel objects, as illustrated in \cref{fig:fig2}. VisTex-OVLM introduces multi-scale textualizing blocks (MSTB) and a multi-stage fusion strategy to achieve the visual textualization of support images. The MSTB maps the visual features of support images into the text feature space, enabling direct support image prompting of an unmodified OVLM model while preserving its excellent pre-trained object-text alignment.

\subsubsection{Visual textualization}
An effective approach to modifying VLM's prediction preferences without altering its weights and architecture is to adjust the text prompts in the language branch. Thus, we propose mapping support images to the text feature space, i.e. visual textualization, to incorporate information about novel classes while preserving OVLM's object-text alignment.

\textbf{Image prompt engineering} Since object detection is a dense prediction task, it is crucial to maximize the utilization of few-shot annotations for distinguishing the target object in the support image from irrelevant objects and the background. Following Luddecke et al.’s \cite{luddecke2022clipseg} experience with image prompting in semantic segmentation, we implement image prompt engineering by combining the support images with the target bounding boxes. Specifically, we apply background blur to regions outside the target bounding box. 

\textbf{Multi-scale textualizing} Let $x_S$ denote the prompt-engineered one-shot support image. We utilize OVLM’s frozen visual encoder to extract multi-scale features containing information about objects of different sizes. For stage $i$, we denote the extracted visual features of $x_S$ as: $R_S^i=\left[ R_S^{i,(j)}\right]_{j=0}^{M-1}$, where $R_S^{i,(j)}\in \mathbb{R}^{\frac{H}{2^j}\cdot \frac{W}{2^j} \times d_I}$. We design a multi-scale textualizing block (MSTB) to process visual features at all scales uniformly and map them to the text feature space, as illustrated in \cref{fig:fig2}. Specifically, we employ $3\times3$ convolutions with stride $2$ to progressively downsample and extract the most effective visual information from larger-scale features, followed by an MLP that maps the visual features to the text feature dimension $d_T$. %=\left[R_S^{i,(0)}, R_S^{i,(1)}, R_S^{i,(2)}, R_S^{i,(3)}, R_S^{i,(4)}\right]

To reduce the cost of training MSTB on the base set $D_b$ and effectively utilize shared knowledge across features of different scales, we adopt a parameter-sharing strategy. Specifically, we apply a shared set of convolutions to process all features except those at the smallest scale, followed by an MLP mapping to obtain the textualized visual feature $\widetilde{P_S^i}$:
\begin{equation}
\begin{aligned}
    \widetilde{R^{i,(j)}_S} = \left\{
    \begin{array}{ll}
    \left( \prod_{j}^{M-2} \operatorname{Conv}_{\text{down}}^{i,(j)}\right) \left( R^{i,(j)}_S \right), & \text{if } j \neq M-1 \\
    R^{i,(j)}_S, & \text{if } j = M-1
    \end{array}
    \right.
\end{aligned}
\label{eq1}
\end{equation}
\begin{equation}
    \widetilde{R^i_S} = \left[ \widetilde{R^{i,(j)}_S} \right]_{j=0}^{M-1} \in \mathbb{R}^{M\cdot \frac{H}{2^{M-1}} \cdot \frac{W}{2^{M-1}} \times d_I},
    \label{eq2}
\end{equation}
\begin{equation}
    \widetilde{P^i_S} = \operatorname{MLP}^i \left( \widetilde{R^i_S} \right) \in \mathbb{R}^{1 \times d_T}.
    \label{eq3}
\end{equation}
This strategy facilitates learning the mapping functions that project different-scale features into a unified text feature space. For FSOD, multi-scale textualizing enables comprehensive feature extraction from support samples across different scales, capturing both fine-grained local details and global contextual relationships. This multi-scale approach significantly enhances the representational capability of $\widetilde{P_S^i}$ for target objects in query images.

\textbf{Multi-stage fusion} We further implement a multi-stage fusion (MSF) strategy to integrate textualized visual features $\widetilde{P_S^i}$ from different visual encoder stages of support images. This fusion process consolidates each support image into a single textualized visual token $\widetilde{P_S}$:
\begin{equation}
    \widetilde{P_S} = \operatorname{MSF}^i \left( \{ \widetilde{R^i_S}\}_{i=0}^L \right).
    \label{eq4}
\end{equation}
This fusion process can be implemented through non-parametric operations, conserving computational resources while achieving effective results. In fact, the synergistic effect of multi-stage fusion and multi-scale textualizing fully leverages OVLM's well-established object-text alignment and object-level visual feature extraction capabilities, enabling the textualized visual token $\widetilde{R^i_S}$ to achieve optimal semantic representation.
\subsubsection{Direct support image prompting}
%Previous attempts to use images for prompting either introduced additional components that potentially compromised OVLM's object-text alignment or merely modulated the text features 
Previous methods that use images for prompting either add extra components \cite{luddecke2022clipseg, xu2023mqdet} that risk disrupting OVLM's object-text alignment or modulate text features \cite{xu2023mqdet} without directly incorporating visual information. In contrast, VisTex-OVLM achieves direct integration of support image visual information through visual textualization.

%We concatenate the textualized visual token $\widetilde{P_S}$ with text prompt tokens tokenized by a pre-trained BERT and directly input them into an unmodified pre-trained OVLM. Specifically, the extraction process for $P^0$ in VisTex-OVLM becomes:
In VisTex-OVLM, we concatenate the textualized visual token $\widetilde{P_S}$ with text prompt tokens processed by a pre-trained BERT and input them directly into an unmodified pre-trained OVLM:
\begin{equation}
    P^0 = \left[ \operatorname{BERT}\left(t\right), \widetilde{P_S} \right].
    \label{eq5}
\end{equation}
% For $K$-shot object detection tasks, to maintain the specific independence of each shot—which positively influences OVLM's ability to capture support sample distribution for query prediction—we directly concatenate textualized visual tokens from different support images rather than applying potentially information-losing fusion operations:
For $K$-shot tasks, we maintain each shot’s independence —-- essential for OVLM to capture support sample distribution for query prediction —-- by directly concatenating textualized tokens from different support images, avoiding information-losing fusion:
\begin{equation}
    P^0 = \left[ \operatorname{BERT}\left(t\right), \widetilde{P_{S_1}}, \cdots, \widetilde{P_{S_K}} \right].
    \label{eq6}
\end{equation}
For a setting with $\lvert C\rvert$ classes, $P^0\in \mathbb{R}^{\lvert C\rvert \cdot \left(N+K\right)\times d_{T}}$. During the training phase, we construct corresponding text prompts $t$ using class labels from the base set $D_b$. Following previous FSOD work, we randomly select $K$-shot samples as support images for each base class, utilizing the remaining data as queries, and train MSTB end-to-end using OVLM's pre-training loss functions. After training, MSTB can directly perform textualization on novel class data.

Unlike previous image prompting methods, we directly embed the semantics of few-shot support images into OVLM's text prompt $P^0$ in the form of textualized visual tokens, enabling direct incorporation of visual information. Our approach maintains the OVLM's original architecture, fully preserving its excellent pre-trained object-text alignment. Implementations of VisTex-OVLM on GLIP and GroundingDINO, namely VisTex-GLIP and VisTex-DINO, demonstrate superior generalization ability and performance. 

\begin{table*}[t]
    \centering
    \resizebox{\textwidth}{!}{
    \begin{tabular}{c|cc|ccccccccccc|ccccc}
    \hline
    \multirow{2}{*}{\textbf{Method}} & \multicolumn{2}{c|}{\textbf{LVIS MiniVal}} & \multicolumn{11}{c|}{\textbf{Unseen subsets from ODinW35}} & \multicolumn{5}{c}{\textbf{Unseen medical datasets}} \\ \cline{2-19} 
     & \textbf{AP} & \textbf{APr} & \textbf{A} & \textbf{B} & \textbf{C} & \textbf{D} & \textbf{E} & \textbf{F} & \textbf{G} & \textbf{H} & \textbf{I} & \textbf{J} & \textbf{K} & \textbf{MoNu} & \textbf{CCRCC} & \textbf{ConSeP} & \textbf{LIDC} & \textbf{Deeplesion} \\ \hline
    \textbf{Meta-DETR \cite{zhang2022metaDETR1}} & 24.6 & 20.1 & 22.8 & 21.5 & 30.5 & 23.6 & 28.9 & 29.4 & 23.6 & 32.4 & 12.2 & 23.1 & 7.3 & 15.5 & 18.2 & 19.9 & 1.2 & 0.6 \\
    \textbf{DiGeo \cite{ma2023digeo}} & 24.1 & 20.5 & 35.9 & 22.6 & 21.4 & 25.1 & 31.5 & 38.7 & 29.6 & 43.1 & 23.1 & 21.8 & 11.6 & 18.2 & 20.4 & 22.8 & 6.1 & 2.3 \\
    \textbf{DeFRCN \cite{Qiao_2021_defrcn2}} & 32.3 & 28.7 & 42.3 & 41.5 & 31.1 & 33.8 & 36.8 & 41.1 & 32.6 & 40.2 & 33.9 & 25.2 & 19.3 & 16.3 & 15.6 & 13.1 & 5.2 & 2.8 \\
    \textbf{MFD \cite{wu2022mfd2}} & 28.9 & 26.4 & 36.2 & 33.5 & 32.3 & 24.8 & 27.2 & 38.5 & 28.4 & 34 & 31.7 & 42.2 & 22.8 & 9.4 & 13.5 & 12.1 & 7.5 & 3.5 \\
    \textbf{MQ-Det \cite{xu2023mqdet}} & 36.2 & 31.1 & 44.8 & 42.9 & 38.1 & 33.5 & 41.8 & 40.1 & 22.6 & 32.1 & 22.4 & 46.3 & 26.1 & 7.1 & 7.6 & 8.2 & 6.3 & 3.1\\
    \textbf{OWL-ViT \cite{minderer2022simple}} & -    & -    & 53.6 & 42.2 & 42.7 & 43.6 & 40.2 & 38.4 & 21.9 & 38.5 & 30.5 & 36.1 & 23.5   & 0.2 & 16.7 & 15.9 & 0.2 & 0.0  \\
    \textbf{OWL-ViTv2 \cite{minderer2023scaling}} & 47.2 & 37.8 & 65.4 & 52.1 & 45.8 & 47.8 & 42.3 & 50.6 & 38.2 & 45.1 & 38.5 & 49.8 & 30.7 &  4.4 & 23.4 & 22.5 & 6.5 & 1.2  \\
    % \textbf{FIBER-ZS} & 35.8 & 29.5 & 2.1 & 0.9 & 0.4 & 0.6 & 0.5 & 0.9 & 0.5 & 1.2 & 1.7 & 0.6 & 0.3 & 1.3 & 0.1 & 0.3 & 0.3 & 0.6 \\
    % \textbf{FIBER-FF*} & 48.4 & 38.8 & 95.7 & 69.2 & 60.5 & 68.8 & 65.2 & 71.7 & 44.6 & 51.7 & 38.9 & 64.2 & 29.4 & 25.5 & 29.5 & 34.7 & 9.2 & 9.8  \\
    % \textbf{VisTex-FIBER} & 49.6 & 41.6 & 79.2 & 69.5 & 61.5 & 66.5 & 63.7 & 72.6 & 50.9 & 55.4 & 40.3 & 60.8 & 31.7 & 26.8 & 32.9 & \textbf{36.4} & 10.5 & 11.4 \\
    \textbf{GroundingDINO-ZS} & 36.1 & 30.8 & 0.1 & 0.1 & 0.3 & 0.3 & 0.2 & 0.8 & 0.4 & 0.7 & 1.3 & 1.3 & 0.3 & 2.0 & 16.2 & 0.7 & 0.1 & 0.0 \\
    \textbf{GroundingDINO-FF*} & 38.3 & 33.4 & 17.5 & 3.6 & 6.9 & 8.2 & 19.7 & 7.2 & 8.1 & 20.9 & 4.2 & 66.9 & 24.4 & 25.8 & \textbf{44.6} & 22 & 2.8 & 0.5  \\
    \textbf{VisTex-DINO (Ours)} & 42.8 & 37.2 & 70.2 & 68.4 & 58.5 & 66.5 & 62.0 & 73.1 & 56.1 & 63.3 & 39.8 & 71.6 & 32.5 & 27.7 & 38.5 & 29.6 & 9.6 & 10.8 \\
    \textbf{GLIP-ZS} & 37.3 & 28.2 & 0.0 & 0.0 & 0.1 & 0.4 & 0.9 & 1.0 & 0.7 & 0.1 & 1.3 & 0.4 & 0.4 & 4.2 & 8.8 & 1.3 & 0.3 & 0.0 \\
    \textbf{GLIP-FF*} & 49.5 & 40.5 & 75.8 & 69.8 & 61.4 & \textbf{70.5} & 67.4 & 75.4 & 55.3 & 63.1 & 44.5 & 68.7 & \textbf{41.2} & 28.5 & 30.6 & 35.1 & 11.3 & \textbf{11.9} \\
    \textbf{VisTex-GLIP (Ours)} & \textbf{50.7} & \textbf{42.9} & \textbf{77.5} & \textbf{70.2} & \textbf{63.1} & 68.0 & \textbf{69.5} & \textbf{75.7} & \textbf{58.4} & \textbf{65.7} & \textbf{46.9} & \textbf{73.5} & 38.9 & \textbf{28.9} & 31.5 & \textbf{35.7} & \textbf{12.2} & 11.7 \\ \hline
    \end{tabular}
    }
    \caption{Performance in open-set scenarios. Subsets A\textasciitilde K from ODinW35 \cite{li2022odinw}: PKLot\_640, openPoetryV, boggleBoards, dicemedCol, OxfPetsbybreed, UnoCards, plantdoc, EgoHands\_s, webScreenshots, OxfPetsbyspecies, MaskWearing. Best results are marked in \textbf{bold}.}
    % \vskip -0.08in
    \label{tab:transfer}
\end{table*}

\begin{table*}[t]
    \centering
    \resizebox{\textwidth}{!}{
    \begin{tabular}{cccccccccccccccccc}
    \hline
    \multicolumn{2}{c|}{} & \multicolumn{5}{c|}{\textbf{Split 1}} & \multicolumn{5}{c|}{\textbf{Split 2}} & \multicolumn{5}{c|}{\textbf{Split 3}} &  \\ \cline{3-17}
    \multicolumn{2}{c|}{\multirow{-2}{*}{\textbf{Method/Shot}}} & \textbf{1} & \textbf{2} & \textbf{3} & \textbf{5} & \multicolumn{1}{c|}{\textbf{10}} & \textbf{1} & \textbf{2} & \textbf{3} & \textbf{5} & \multicolumn{1}{c|}{\textbf{10}} & \textbf{1} & \textbf{2} & \textbf{3} & \textbf{5} & \multicolumn{1}{c|}{\textbf{10}} & \multirow{-2}{*}{\textbf{Mean}} \\ \hline
    \multicolumn{18}{c}{\textbf{non-VLM-based}} \\ \hline
    \textbf{MetaDet \cite{wang2019metadet}} & \multicolumn{1}{c|}{\textbf{ICCV 19}} & 18.9 & 20.6 & 30.2 & 36.8 & \multicolumn{1}{c|}{49.6} & 21.8 & 23.1 & 27.8 & 31.7 & \multicolumn{1}{c|}{43.0} & 20.6 & 23.9 & 29.4 & 43.9 & \multicolumn{1}{c|}{44.1} & 31.0 \\
    \textbf{MPSR \cite{wu2020mpsr}} & \multicolumn{1}{c|}{\textbf{ECCV 20}} & 41.7 & - & 51.4 & 55.2 & \multicolumn{1}{c|}{61.8} & 24.4 & - & 39.2 & 39.9 & \multicolumn{1}{c|}{47.8} & 35.6 & - & 42.3 & 48.0 & \multicolumn{1}{c|}{49.7} & 44.8 \\
    \textbf{DeFRCN \cite{Qiao_2021_defrcn2}} & \multicolumn{1}{c|}{\textbf{ICCV 21}} & 53.6 & 57.5 & 61.5 & 64.1 & \multicolumn{1}{c|}{60.8} & 30.1 & 38.1 & 47.0 & 53.3 & \multicolumn{1}{c|}{47.9} & 48.4 & 50.9 & 52.3 & 54.9 & \multicolumn{1}{c|}{57.4} & 51.9 \\
    \textbf{Meta-DETR \cite{zhang2022metaDETR1}} & \multicolumn{1}{c|}{\textbf{TPAMI 22}} & 40.6 & 51.4 & 58.0 & 59.2 & \multicolumn{1}{c|}{63.6} & 37.0 & 36.6 & 43.7 & 49.1 & \multicolumn{1}{c|}{54.6} & 41.6 & 45.9 & 52.7 & 58.9 & \multicolumn{1}{c|}{60.6} & 50.2 \\
    \textbf{MFD \cite{wu2022mfd2}} & \multicolumn{1}{c|}{\textbf{ECCV 22}} & 63.4 & 66.3 & 67.7 & 69.4 & \multicolumn{1}{c|}{68.1} & 42.1 & 46.5 & 53.4 & 55.3 & \multicolumn{1}{c|}{53.8} & 56.1 & 58.3 & 59.0 & 62.2 & \multicolumn{1}{c|}{63.7} & 59.0 \\
    \textbf{ICPE \cite{lu2023ICPE}} & \multicolumn{1}{c|}{\textbf{AAAI 23}} & 54.3 & 59.5 & 62.4 & 65.7 & \multicolumn{1}{c|}{66.2} & 33.5 & 40.1 & 48.7 & 51.7 & \multicolumn{1}{c|}{52.5} & 50.9 & 63.1 & 55.3 & 60.6 & \multicolumn{1}{c|}{60.1} & 55.0 \\
    \textbf{VFA \cite{han2023vfa}} & \multicolumn{1}{c|}{\textbf{AAAI 23}} & 57.7 & 64.6 & 64.7 & 67.2 & \multicolumn{1}{c|}{67.4} & 41.4 & 46.2 & 51.1 & 51.8 & \multicolumn{1}{c|}{51.6} & 48.9 & 54.8 & 56.6 & 59.0 & \multicolumn{1}{c|}{58.9} & 56.1 \\
    \textbf{Du et al. \cite{Du_2023_ICCV}} & \multicolumn{1}{c|}{\textbf{ICCV 23}} & 52.3 & 55.5 & 63.1 & 65.9 & \multicolumn{1}{c|}{66.7} & 42.7 & 45.8 & 48.7 & 54.8 & \multicolumn{1}{c|}{56.3} & 47.8 & 51.8 & 56.8 & 60.3 & \multicolumn{1}{c|}{62.4} & 55.4 \\
    \textbf{SNIDA-MFD \cite{wang2024snida2}} & \multicolumn{1}{c|}{\textbf{CVPR 24}} & 64.9 & 67.9 & 69.7 & 71.4 & \multicolumn{1}{c|}{70.5} & 42.2 & 47.8 & 54.5 & 56.6 & \multicolumn{1}{c|}{54.9} & 58.1 & 61.3 & 60.7 & 63.6 & \multicolumn{1}{c|}{66.0} & 60.7 \\ \hline
    \multicolumn{18}{c}{\textbf{VLM-based}} \\ \hline
    \textbf{D\&R \cite{li2023dr}} & \multicolumn{1}{c|}{\textbf{AAAI 23}} & 41.0 & 51.7 & 55.7 & 61.8 & \multicolumn{1}{c|}{65.4} & 30.7 & 39.0 & 42.5 & 46.6 & \multicolumn{1}{c|}{51.7} & 37.9 & 47.1 & 51.7 & 56.8 & \multicolumn{1}{c|}{59.5} & 49.3 \\
    \textbf{Norm-VAE \cite{xu2023normvae}} & \multicolumn{1}{c|}{\textbf{CVPR 23}} & 62.1 & 64.9 & 67.8 & 69.2 & \multicolumn{1}{c|}{67.5} & 39.9 & 46.8 & 54.4 & 54.2 & \multicolumn{1}{c|}{53.6} & 58.2 & 60.3 & 61.0 & 64.0 & \multicolumn{1}{c|}{65.5} & 59.1 \\
    \textbf{FM-FSOD-L \cite{Han_2024_FM-FSOD}} & \multicolumn{1}{c|}{\textbf{CVPR 24}} & 40.1 & 53.5 & 57.0 & 68.6 & \multicolumn{1}{c|}{72.0} & 33.1 & 36.3 & 48.8 & 54.8 & \multicolumn{1}{c|}{64.7} & 39.2 & 50.2 & 55.7 & 63.4 & \multicolumn{1}{c|}{68.1} & 53.7 \\
    \textbf{DP-DDCL \cite{guo2024DP-DDCL}} & \multicolumn{1}{c|}{\textbf{KBS 24}} & 49.4 & 60.1 & 63.9 & 67.5 & \multicolumn{1}{c|}{69.1} & 37.5 & 43.4 & 48.4 & 52.4 & \multicolumn{1}{c|}{56.2} & 45.4 & 56.3 & 59.0 & 63.0 & \multicolumn{1}{c|}{65.7} & 55.8 \\
    \textbf{VEIC \cite{zhao2024veic}} & \multicolumn{1}{c|}{\textbf{ESWA 24}} & 67.7 & 69.5 & 70.5 & 70.4 & \multicolumn{1}{c|}{69.1} & 49.5 & 51.5 & 55.0 & 56.2 & \multicolumn{1}{c|}{54.7} & 63.3 & 64.7 & 63.6 & 65.4 & \multicolumn{1}{c|}{65.4} & 62.4 \\
    \textbf{MTL-FSOD \cite{ren2024MTL-FSOD}} & \multicolumn{1}{c|}{\textbf{ECCV 24}} & 68.9 & 71.5 & 72.1 & 74.5 & \multicolumn{1}{c|}{72.2} & 65.5 & 69.8 & \textbf{73.5} & \textbf{74.4} & \multicolumn{1}{c|}{73.1} & 68.8 & 69.8 & 70.0 & 71.6 & \multicolumn{1}{c|}{71.9} & 71.2 \\

    \textbf{OWL-ViT \cite{minderer2022simple}} & \multicolumn{1}{c|}{\textbf{ECCV 22}} & 45.3 & 48.2 & 51.4 & 52.3 & \multicolumn{1}{c|}{51.7} & 42.8 & 43.1 & 48.2 & 51.2 & \multicolumn{1}{c|}{50.3} & 43.9 & 45.6 & 47.3 & 49.1 & \multicolumn{1}{c|}{49.6} & 48.0 \\
    \textbf{OWL-ViTv2 \cite{minderer2023scaling}} & \multicolumn{1}{c|}{\textbf{NeurlPS 23}} & 48.6 & 49.3 & 51.2 & 52.6 & \multicolumn{1}{c|}{53.1} & 45.7 & 47.8 & 50.2 & 52.1 & \multicolumn{1}{c|}{51.4} & 42.6 & 46.2 & 49.6 & 51.0 & \multicolumn{1}{c|}{51.8} & 49.5 \\
    
    \textbf{MQ-Det \cite{xu2023mqdet}} & \multicolumn{1}{c|}{\textbf{NeurlPS 23}} & 53.9 & 58.4 & 61.4 & 64.7 & \multicolumn{1}{c|}{65.1} & 49.2 & 51.3 & 55.6 & 58.4 & \multicolumn{1}{c|}{60.2} & 49.9 & 53.5 & 57.4 & 60.2 & \multicolumn{1}{c|}{63.1} & 57.5 \\
     
    \textbf{GroundingDINO-ZS} & \multicolumn{1}{c|}{\textbf{ECCV 24}} & \multicolumn{5}{c|}{------------------49.6------------------} & \multicolumn{5}{c|}{------------------43.3------------------} & \multicolumn{5}{c|}{------------------47.4------------------} & 46.8  \\
     
    \textbf{GroundingDINO-FF} & \multicolumn{1}{c|}{\textbf{ECCV 24}} & 49.5 & 50.4 & 51.1 & 52.7 & \multicolumn{1}{c|}{53.1} & 42.8 & 43.6 & 44.8 & 46.5 & \multicolumn{1}{c|}{47.8} & 47.5 & 48.9 & 50.8 & 52.4 & \multicolumn{1}{c|}{52.8} & 49.0 \\
     
    \textbf{VisTex-DINO} & \multicolumn{1}{c|}{\textbf{Ours}} & 51.8 & 53.1 & 55.8 & 57 & \multicolumn{1}{c|}{60.2} & 43.5 & 46.7 & 48.5 & 51.1 & \multicolumn{1}{c|}{53.2} & 48.6 & 50.6 & 53.9 & 56 & \multicolumn{1}{c|}{58.5} & 52.6 \\
     
    \textbf{GLIP-ZS} & \multicolumn{1}{c|}{\textbf{CVPR 22}} & \multicolumn{5}{c|}{------------------50.5------------------} & \multicolumn{5}{c|}{------------------43.0------------------} & \multicolumn{5}{c|}{------------------49.5------------------} & 47.7 \\
     
    \textbf{GLIP-FF} & \multicolumn{1}{c|}{\textbf{CVPR 22}} & 51.3 & 55.2 & 59.4 & 61.2 & \multicolumn{1}{c|}{61.8} & 44.3 & 46.2 & 48.6 & 49.3 & \multicolumn{1}{c|}{51.2} & 50.1 & 52.6 & 58.3 & 61.4 & \multicolumn{1}{c|}{61.9} & 54.2 \\
     
    \textbf{VisTex-GLIP} & \multicolumn{1}{c|}{\textbf{Ours}} & \textbf{69.4} & \textbf{71.8} & \textbf{73.4} & \textbf{74.6} & \multicolumn{1}{c|}{\textbf{74.8}} & \textbf{66.5} & \textbf{70.1} & 73.1 & 74.2 & \multicolumn{1}{c|}{\textbf{74.4}} & \textbf{68.9} & \textbf{70.2} & \textbf{71.1} & \textbf{72.3} & \multicolumn{1}{c|}{\textbf{72.6}} & \textbf{71.8} \\ 
     \hline
    \end{tabular}
    }
    \caption{Comparison of different FSOD methods in terms of AP50 on three PASCAL VOC Novel Split sets. Best results are marked in \textbf{bold}.}
    % \vskip -0.1in
    \label{tab:pascal-com}
\end{table*}

\section{Experiment}
\subsection{Datasets}
\textcolor{black}{\textbf{Open-set scenarios:} To rigorously validate VisTex-OVLM’s effectiveness on genuinely novel categories, we evaluated its transferability in open-set scenarios, including LVIS MiniVal \cite{gupta2019lvis} and 16 unseen datasets with minimal overlap with OVLM pre-training data. LVIS challenges long-tail object detection, while 11 datasets from ODinW35 \cite{li2022odinw} (mAP$\leq2$ for GLIP-L) and 5 medical datasets (MoNu \cite{kumar2017datasetMONU}, CCRCC \cite{gao2021nuclei}, ConSep \cite{graham2019hover}, LIDC \cite{armato2004lungLIDC}, and Deeplesion \cite{yan2017deeplesion}) test robustness in non-natural imaging contexts.}

\textbf{Standard FSOD benchmarks} Following previous FSOD works \cite{wang2020frustratingly, zhang2022metaDETR1}, we evaluated our method on PASCAL VOC \cite{everingham2010pascalvoc} and MSCOCO \cite{lin2014microsoftCOCO}. \textbf{PASCAL VOC:} We used three partitions of base and novel categories (SPLIT1, SPLIT2, and SPLIT3). Each partition divides the 20 PASCAL VOC categories into 15 base classes ($C_b$) and 5 novel classes ($C_n$), reporting AP50 results on the novel set ($D_n$) under 1, 2, 3, 5, and 10 shots. \textbf{MSCOCO:} For MSCOCO's 80 classes, the 20 classes overlapping with PASCAL VOC are designated as novel ($C_n$), and the remaining 60 as base ($C_b$), with mAP results reported under 1, 2, 3, 5, 10, and 30 shots.  % Images from the base classes are used during training, while in testing, both the original and support images come from the novel classes.

% \textbf{Real-world downstream tasks:} Without loss of generality, we also assessed VisTex-OVLM’s transferability based on GLIP, across 16 real-world downstream datasets. ODinW13 \cite{li2022odinw} spans 13 specialized natural domains, such as aquarium species, surgical instruments, and aerial imagery, etc. Additionally, we included 3 medical datasets --— MoNu \cite{kumar2017datasetMONU}, a nuclei detection dataset from H\&E stained pathology, and two CT lesion detection datasets, LIDC \cite{armato2004lungLIDC} and Deeplesion \cite{yan2017deeplesion}—to test VisTex-OVLM’s performance on non-natural scenes.

% \textbf{Real-world downstream tasks:} Beyond these two benchmarks, we further evaluated VisTex-GLIP's transferability on 16 challenging real-world downstream task datasets. ODinW13 \cite{li2022odinw} covers 13 specialized scenes and professional domain objects, such as aquarium species, surgical instruments, aerial imagery, etc. Additionally, since GLIP is typically effective in recognizing objects in natural scenes but its ability to identify unseen objects in non-natural scenes remains uncertain, we incorporate 3 medical image datasets for further evaluation, including Monu\cite{kumar2017datasetMONU}, LIDC\cite{armato2004lungLIDC}, and Deeplesion\cite{yan2017deeplesion}. Monu is an H\&E stained pathology dataset for nuclei detection while LIDC and Deeplesion are CT datasets for lesion detection.

\subsection{Implementation details}

Without loss of generality, we conducted main experiments using the pre-trained GLIP-L \cite{li2022glip}, referred to as VisTex-GLIP. We also implemented VisTex-OVLM on GroundingDINO-T \cite{liu2023groundingdino}, denoted as VisTex-DINO. Unless specified otherwise, we used the following experimental settings for VisTex-GLIP.

The MSTB's MLP consists of two "fully connected (fc) + ReLU" layers. The two fc layers are responsible for spatial aggregation $\left( M\cdot \frac{H}{2^{M-1}} \cdot \frac{W}{2^{M-1}} \rightarrow 1 \right)$ and channel dimension transformation $\left( d_I \rightarrow d_T \right)$, respectively. According to the feature size and scale extracted by the vision encoder of the pre-trained GLIP-L, both $H$ and $W$ are set to 100, and $M$ is 5. By default, MSTB is applied to all 8 stages of GLIP-L, and max pooling across stages is used for non-parametric multi-stage fusion. % MSTB is added to all eight stages of GLIP-L, and maxpooling on the stage dimension is selected as the non-parametric multi-stage fusion method.

For the training of MSTB, we utilized 2 Nvidia RTX 3090 GPUs, each contains 24 GB of memory. For each epoch on ($D_b$), $K$-shot support images were randomly selected for each class $C \in C_b$, while the rest served as query images. Based on query image labels, the corresponding support images were used to generate textualized image tokens $\widetilde{P_S}$, which were then concatenated with text prompts. The training spanned 30 epochs using AdamW with an initial learning rate of $10^{-4}$ and weight decay of 0.01. %During training on ($D_b$), for each epoch, we randomly selected $K$-shot support images for each class $C \in C_b$, with the rest serving as query images. Based on the category labels of the query images, we selected the corresponding support images to generate textualized image tokens $\widetilde{P_S}$ and concatenated them with text prompt tokens. Each epoch requires iterating through all the query images. We train the model with 30 epoch. The model’s weights and biases are optimized using an AdamW optimizer with an init learning rate of $10^{-4}$  and weight decay of 0.01.

\begin{table}[t]
    \centering
    \resizebox{\linewidth}{!}{
    \begin{tabular}{cccccccc}
    \hline
    \multicolumn{2}{c|}{} & \multicolumn{6}{c}{\textbf{Shot Number}} \\ \cline{3-8} 
    \multicolumn{2}{c|}{\multirow{-2}{*}{\textbf{Method}}} & \textbf{1} & \textbf{2} & \textbf{3} & \textbf{5} & \textbf{10} & \textbf{30} \\ \hline
    \multicolumn{8}{c}{\textbf{non-VLM-based}} \\ \hline
    \textbf{Meta R-CNN \cite{yan2019metarcnn}} & \multicolumn{1}{c|}{\textbf{ICCV 19}} & 1.0 & 1.8 & 2.8 & 4.0 & 6.5 & 11.1 \\
    \textbf{MPSR \cite{wu2020mpsr}} & \multicolumn{1}{c|}{\textbf{ECCV 20}} & 5.1 & 6.7 & 7.4 & 8.7 & 9.8 & 14.1 \\
    \textbf{DeFRCN \cite{Qiao_2021_defrcn2}} & \multicolumn{1}{c|}{\textbf{ICCV21}} & 9.3 & 12.9 & 14.8 & 16.1 & 18.5 & 22.6 \\
    \textbf{Meta-DETR \cite{zhang2022metaDETR1}} & \multicolumn{1}{c|}{\textbf{TPAMI 22}} & 7.5 & - & 13.5 & 15.4 & 19.0 & 22,2 \\
    \textbf{MFD \cite{wu2022mfd2}} & \multicolumn{1}{c|}{\textbf{ECCV 22}} & 10.8 & 13.9 & 15.0 & 16.4 & 19.4 & 22.7 \\
    \textbf{Du et al. \cite{Du_2023_ICCV}} & \multicolumn{1}{c|}{\textbf{ICCV23}} & - & - & - & - & 20.3 & 22.8 \\
    \textbf{SNIDA-MFD \cite{wang2024snida2}} & \multicolumn{1}{c|}{\textbf{CVPR 24}} & 12.0 & 15.4 & 16.4 & 17.8 & 20.7 & 23.8 \\ \hline
    \multicolumn{8}{c}{\textbf{VLM-based}} \\ \hline
    \textbf{D\&R \cite{li2023dr}} & \multicolumn{1}{c|}{\textbf{AAAI23}} & 8.3 & 12.7 & 14.3 & 16.4 & 18.7 & 21.8 \\
    \textbf{Norm-VAE \cite{xu2023normvae}} & \multicolumn{1}{c|}{\textbf{CVPR23}} & 9.5 & 13.7 & 14.3 & 15.9 & 18.7 & 22.5 \\
    \textbf{FM-FSOD-L \cite{Han_2024_FM-FSOD}} & \multicolumn{1}{c|}{\textbf{CVPR 24}} & 5.7 & 11.0 & 15.7 & 21.9 & 27.7 & 37.0 \\
    \textbf{DP-DDCL \cite{guo2024DP-DDCL}} & \multicolumn{1}{c|}{\textbf{KBS 24}} & 9.0 & 12.7 & 14.6 & 17.2 & 19.9 & 23.0 \\
    \textbf{VEIC \cite{zhao2024veic}} & \multicolumn{1}{c|}{\textbf{ESWA 24}} & 11.5 & 14.4 & 15.3 & 16.9 & 19.7 & 21.9 \\
    \textbf{MTL-FSOD \cite{ren2024MTL-FSOD}} & \multicolumn{1}{c|}{\textbf{ECCV 24}} & 12.8 & 16.9 & 17.5 & 19.5 & 22.7 & 25.2 \\
    \textbf{OWL-ViT \cite{minderer2022simple}} & \multicolumn{1}{c|}{\textbf{ECCV 22}} & 22.2 & 26.1 & 31.4 & 32.4 & 31.5 & 32.8 \\
    \textbf{OWL-ViTv2 \cite{minderer2023scaling}} & \multicolumn{1}{c|}{\textbf{NeurlPS 23}} & 26.5 & 28.3 & 30.5 & 32.9 & 33.5 & 33.1 \\
   
    \textbf{MQ-Det \cite{xu2023mqdet}} & \multicolumn{1}{c|}{\textbf{NeurlPS 23}} & 42.2 & 46.3 & 47.5 & 47.7 & 47.8 & 48.1 \\
    
    \textbf{GroundingDINO-ZS} & \multicolumn{1}{c|}
    {\textbf{ECCV 24}} & \multicolumn{6}{c}{---------------------48.5---------------------} \\
    
    \textbf{GroundingDINO-FF} & \multicolumn{1}{c|}{\textbf{ECCV 24}} & 48.7 & 48.9 & 49.3 & 50.8 & 51.6 & 53.0 \\
    
    \textbf{VisTex-DINO} & \multicolumn{1}{c|}{\textbf{Ours}} & \textbf{48.9} & 49.2 & 50.2 & 52.5 & \textbf{54.3} & \textbf{55.4} \\ 
    \textbf{GLIP-ZS} & \multicolumn{1}{c|}{\textbf{CVPR 22}} & \multicolumn{6}{c}{---------------------42.9---------------------} \\
     
    \textbf{GLIP-FF} & \multicolumn{1}{c|}{\textbf{CVPR 22}} & 41.2 & 44.7 & 47.5 & 48.8 & 49.3 & 49.4 \\
     
    \textbf{VisTex-GLIP} & \multicolumn{1}{c|}{\textbf{Ours}} & 47.9 & \textbf{51.8} & \textbf{52.2} & \textbf{52.6} & 52.7 & 53.6 \\ 
     \hline
    \end{tabular}
    }
    \caption{FSOD performance on MSCOCO. Best results are marked in \textbf{bold}.}
    \vskip -0.26in
    \label{tab:MSCOCO-com}
\end{table}

\textcolor{black}{Detailed implementation for VisTex-DINO and comparison methods are provided in the supplementary materials.} %. For methods that reported results on PASCAL VOC and MSCOCO in their original papers, we directly adopted the published values. For those that did not, we reproduced the results on these two benchmarks using their default settings.

\subsection{Open-set scenarios}
\label{sec:4.4}
% We evaluated VisTex-OVLM’s transferability using GLIP on challenging downstream datasets, comparing it with methods with available code and weights, and reporting mAP in \cref{tab:transfer}. All methods, including ours, were trained on MSCOCO’s base set, treating downstream task sets as novel and testing with 2-shot support images. Our method significantly outperformed others on ODinW13 and three medical datasets, demonstrating strong transferability across large domain gaps. However, due to MSCOCO's limited similarity to medical data, direct transfer to medical datasets was initially lower; thus, we briefly fine-tuned MSTB in a 2-shot setting on medical tasks, yielding final mAPs of 38.3, 30.2, and 35.1. From the results above, we can conclude that our method outperforms other methods when directly shifting domains on medical datasets, and can quickly adapt and improve with minor re-fine-tuning. Detailed transfer results for ODinW13 subsets are in the supplementary materials.

\textcolor{black}{We evaluated VisTex-OVLM’s performance in open-set scenarios, including LVIS MiniVal and 16 unseen datasets with categories exhibiting minimal overlap with the OVLM pre-training data. We compared VisTex-OVLM with methods that have publicly available code and weights, reporting mAP unless otherwise specified in \cref{tab:transfer}. We established two baselines: OVLM-ZS and OVLM-FF, e.g., GLIP-ZS for GLIP. In OVLM-ZS, we used pre-trained OVLM weights for zero-shot testing. In OVLM-FF, we fine-tuned all pre-trained parameters on random 2-shot support images from the novel dataset, using class names as text prompts for both training and test evaluation, and thus is denoted by the "*" marker in the results. Notably, OVLM-ZS cannot incorporate support images' visual information during testing. Other methods requiring training, including ours, were trained on MSCOCO’s base set, treating open-set task sets as novel and testing with 2-shot support images, while OWL-ViT (v2) and OVLM-ZS were evaluated using their frozen pre-trained weights. The results represent the average of five experimental runs. The results show that the zero-shot performance of OVLM significantly drops for novel classes or domains rarely encountered during pre-training compared to its performance on PASCAL VOC and MSCOCO. In contrast, our method outperforms or matches other approaches in most cases, demonstrating strong transferability. It is worth noting that OVLM-FF may achieve slightly higher performance in some cases due to the introduction of new textual knowledge through full fine-tuning. However, VisTex-OVLM, without any training on these datasets, still achieves comparable results. These findings underscore VisTex-OVLM’s robust performance in open-set detection and its broad applicability across diverse domains.}

\subsection{Comparison on FSOD benchmarks}

We compared our method with other recent FSOD methods, including both non-VLM-based methods and VLM-based methods, with results presented in \cref{tab:pascal-com,tab:MSCOCO-com}. In OVLM-FF, we fine-tuned all pre-trained parameters on $K$-shot support images from the novel dataset. Additionally, we directly compared our method with MQ-Det, which is also based on OVLM and uses support images to modulate text prompts for image prompting. \cref{tab:pascal-com,tab:MSCOCO-com} demonstrate the superiority of our approach over common transfer methods in utilizing a limited number of support images by textualizing them and directly prompting OVLM, for both GLIP and GroundingDINO. 
% This is because other methods may disrupt the original object-text alignment when introducing visual information, whereas our approach fully preserves OVLM’s generalization, delivering superior performance on novel sets and achieving SOTA results among FSOD methods, surpassing both VLM-based and non-VLM-based approaches. %Taking the performance on PASCAL VOC SPLIT 1 as an example, GLIP-zero-shot achieved 42.4 mAP, and 1-shot fine-tuning raised it to 44.6. MQ-Det only reached 42.2, while our method achieved 64.1, 

Our method achieves SOTA performance in classic FSOD while retaining strong recognition for base categories, meeting generalized FSOD (GFSOD) requirements. \cref{tab:GFSOD} shows comparisons with other GFSOD methods on MSCOCO in 10-shot and 30-shot settings. By preserving and leveraging VLMs' generalization, our method avoids overfitting and knowledge forgetting on base classes, achieving high performance across both base and novel categories.

\textcolor{black}{We also conducted compatibility experiments on RegionCLIP and FIBER in the supplementary materials to further validate the generalization ability of VisTex-OVLM.}

\begin{table}[t]
    \centering
    \resizebox{0.8\linewidth}{!}{
    \begin{tabular}{ccccccc}
    \hline
    \multicolumn{1}{c|}{} & \multicolumn{3}{c|}{\textbf{10-shot}} & \multicolumn{3}{c}{\textbf{30-shot}} \\ \cline{2-7} 
    \multicolumn{1}{c|}{\multirow{-2}{*}{\textbf{Method}}} & \textbf{AP} & \textbf{bAP} & \multicolumn{1}{c|}{\textbf{nAP}} & \textbf{AP} & \textbf{bAP} & \textbf{nAP} \\ \hline
    \multicolumn{7}{c}{\textbf{non-VLM-based}} \\ \hline
    \multicolumn{1}{c|}{\textbf{DiGeo \cite{ma2023digeo}}} & 32.0 & 39.2 & \multicolumn{1}{c|}{10.3} & 33.1 & 39.4 & 14.2 \\
    \multicolumn{1}{c|}{\textbf{DE-VIT-L \cite{zhang2023devit}}} & 30.6 & 29.4 & \multicolumn{1}{c|}{34.0} & 30.6 & 29.5 & 34.0 \\ \hline
    \multicolumn{7}{c}{\textbf{VLM-based}} \\ \hline
    \multicolumn{1}{c|}{\textbf{FM-FSOD-L \cite{Han_2024_FM-FSOD}}} & 40.0 & 44.2 & \multicolumn{1}{c|}{27.7} & 43.1 & 45.2 & 37.0 \\
    \multicolumn{1}{c|}{\textbf{MQ-Det \cite{xu2023mqdet}}} & 46.9 & \textbf{46.5} & \multicolumn{1}{c|}{47.8} & 47.3 & 46.8 & 48.1 \\
     
    \multicolumn{1}{c|}{\textbf{VisTex-GLIP}} & \textbf{48.1} & 46.2 & \multicolumn{1}{c|}{\textbf{52.7}} & \textbf{48.9} & \textbf{47.2} & \textbf{53.6} \\ \hline
    \end{tabular}
    }
    \caption{GFSOD performance on MSCOCO. Best results are marked in \textbf{bold}.}
    % \vskip -0.2in
    \label{tab:GFSOD}
\end{table}

\begin{figure}[t]
    \centering
    \includegraphics[width=\linewidth]{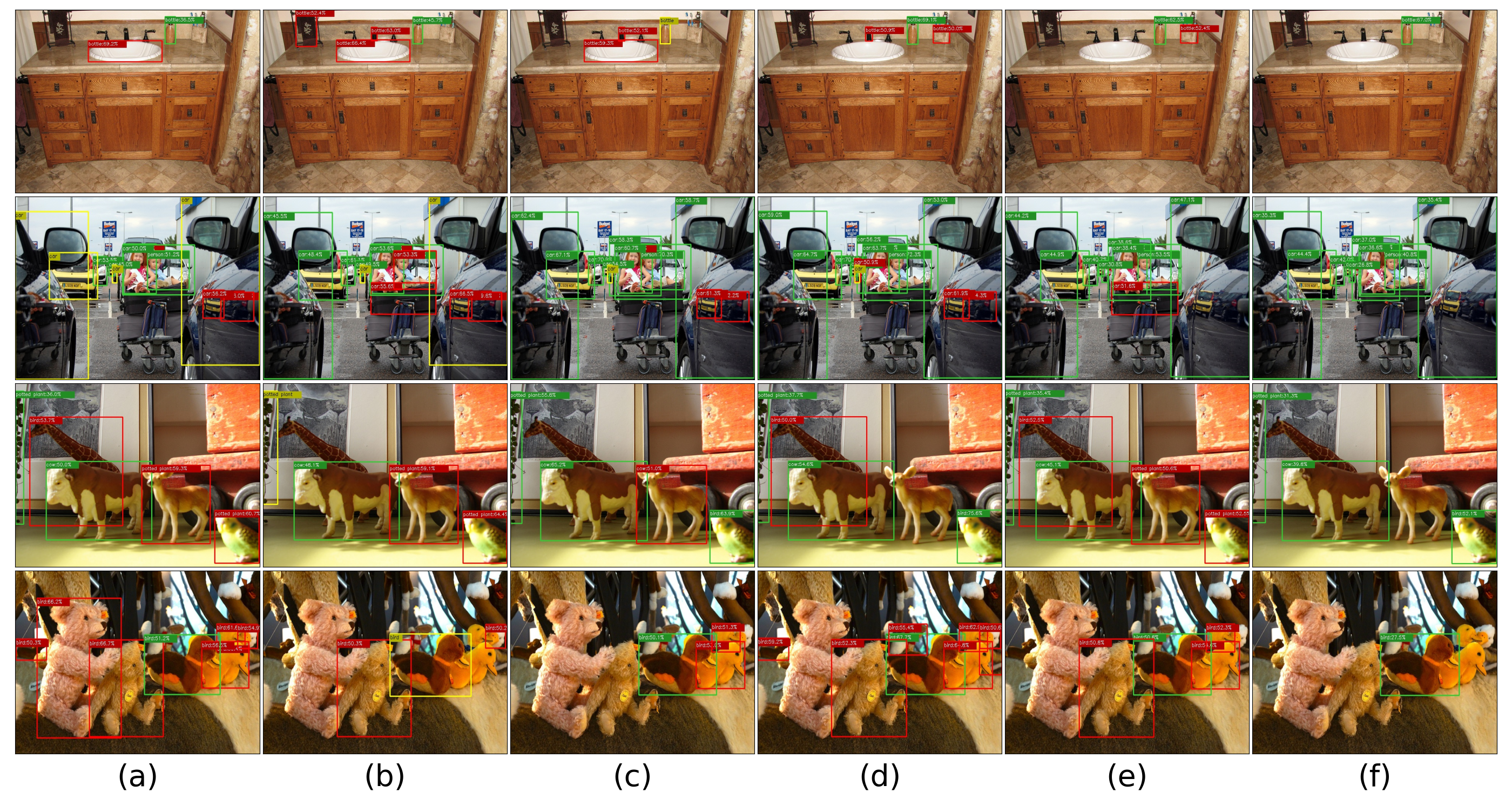}
    % \vskip -0.08in
    \caption{Comparison output visualizations on COCO. (a) Meta-DETR, (b) DiGeo, (c) DeFRCN, (d) MFD, (e) MQ-Det, (f) VisTex-GLIP. Green, red, and yellow boxes denote true positives, false positives, and false negatives.}
    \vskip -0.3in
    \label{fig:vis_comparison}
\end{figure}

\subsection{Ablation study}
We conducted extensive ablation studies on VisTex-OVLM in a 2-shot setting on MSCOCO using GLIP-L, reporting mAP on both the base and the novel classes as in \cref{tab:GFSOD}. All other settings were kept at optimal configurations. More ablation results and analysis regarding VisTex-OVLM's computational overhead are in the supplementary materials.

% \subsubsection{}
We examined the effects of the prompting mode and crucial components, as shown in \cref{tab:ab-prompting}. Text prompts were generated by concatenating class words following the GLIP paper \cite{li2022glip}. We compared GLIP-ZS, GLIP-FF, and GLIP-MaPLe (integrating MaPLe \cite{khattak2023pt}) as a prompt-tuning example.  Multi-scale textualization is fundamental to our method, so the "MSTB sharing" column shows only whether parameter-sharing is applied in MSTB. "$\mathsf{x}$" in the "MSF" column for "VisTex-GLIP" indicates that textualization is applied only in stage 1. As detailed in \cref{tab:ab-prompting}, combining components achieves the best mAP for VisTex-GLIP. Utilizing visual support images addresses GLIP-ZS's limitation of relying solely on text prompts. VisTex-GLIP outperforms GLIP-FF, GLIP-MaPLe, and MQ-Det, which disrupt the object-text alignment by modifying model weights or structure. In contrast, MSTB avoids fine-tuning on novel classes, directly mapping novel support samples into textualized image tokens for prompting, which is user-friendly and avoids disrupting the object-text alignment of GLIP. Notably, image prompts alone without text prompts reduce performance, as text prompts offer essential semantic guidance, impacting VLMs’ image-text alignment. 

\begin{figure}[t]
    \centering
    \includegraphics[width=\linewidth]{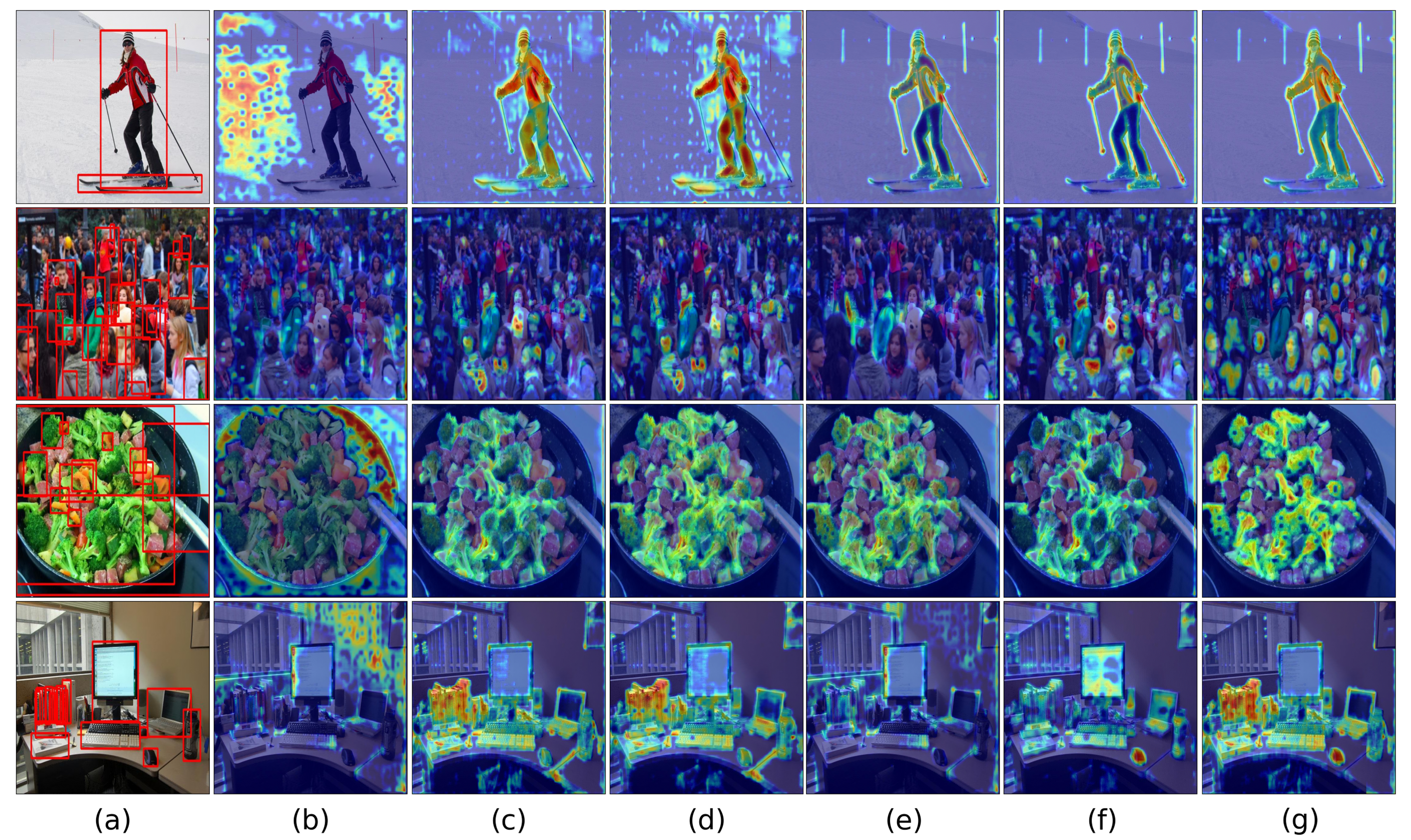}
    \caption{Attention heatmaps comparison. (a) Ground truth, (b) GLIP-ZS (zero-shot), (c) GLIP-FF (fully fine-tuning), (d) GLIP-MaPLe, (e) MQ-Det, (f) VisTex-GLIP without text prompts, (g) VisTex-GLIP.}
    \vskip -0.18in
    \label{fig:heatmap}
\end{figure}

\textcolor{black}{We present attention heatmaps (\cref{fig:heatmap}) showing the effects of different prompting modes and key components, aligned with methods in \cref{tab:ab-prompting}. These heatmaps visualize object feature attention relative to text features. For "VisTex-GLIP without text prompts", the heatmap represents attention exclusively to textualized visual tokens. The heatmaps show that GLIP-FF and GLIP-MAPLE exhibit significant background noise, while MQ-Det suppresses noise but shows less focused object attention. In contrast, VisTex-GLIP achieves stronger target focus and reduced noise. Notably, VisTex-GLIP without text prompts shows weaker target attention than with text prompts. This observation underscores the importance of text prompts in providing essential semantic guidance, which crucially influences OVLMs’ object-text alignment and overall performance.}
 % highlighting the importance of text guidance in shaping OVLMs’ object-text alignment and performance.

% Supplementary materials include attention heatmap comparisons for this study. % Furthermore, both GLIP-MaPLe and MQ-Det require fine-tuning of new components on new datasets, a process that greatly impacts the application flexibility and deployment speed of these methods. %Since GLIP-FF and GLIP-MaPLe are directly fine-tuned with support images, the most direct way to introduce visual information, they are marked with "$\checkmark^{*}$" in the "image prompt" column.

\begin{table}[t]
    \centering
    \resizebox{0.96\linewidth}{!}{
    \begin{tabular}{c|cccc|c|ccc}
    \hline
    \multirow{2}{*}{\textbf{Method}} & \multirow{2}{*}{\textbf{\begin{tabular}[c]{@{}c@{}}Text\\ Prompt\end{tabular}}} & \multirow{2}{*}{\textbf{\begin{tabular}[c]{@{}c@{}}Image\\ Prompt\end{tabular}}} & \multirow{2}{*}{\textbf{\begin{tabular}[c]{@{}c@{}}MSTB\\ sharing\end{tabular}}} & \multirow{2}{*}{\textbf{MSF}} & \multirow{2}{*}{\textbf{\#Par(M)}} & \multicolumn{3}{c}{\textbf{2-shot}} \\ \cline{7-9} 
     &  &  &  &  &  & \textbf{AP} & \textbf{bAP} & \textbf{nAP} \\ \hline
    \textbf{GLIP-ZS} & \textbf{\checkmark} & \textbf{×} & \textbf{×} & \textbf{×} & 0.00 & 42.9 & 44.9 & 40.5 \\
    \textbf{GLIP-FF} & \textbf{\checkmark} & \textbf{\checkmark*} & \textbf{×} & \textbf{×} & 397.59 & 44.3 & 44.1 & 44.7 \\
    \textbf{GLIP-MaPLe} & \textbf{\checkmark} & \textbf{\checkmark*} & \textbf{×} & \textbf{×} & 5.92 & 48.5 & 47.8 & 49.7 \\
    \textbf{MQ-Det} & \textbf{\checkmark} & \textbf{\checkmark} & \textbf{×} & \textbf{×} & 53.10 & 44.7 & 43.5 & 46.3 \\ \hline
    \multirow{4}{*}{\textbf{VisTex-GLIP}} & \textbf{×} & \textbf{\checkmark} & \textbf{×} & \textbf{×} & 38.52 & 30.6 & 32.5 & 25.9 \\
     & \textbf{\checkmark} & \textbf{\checkmark} & \textbf{×} & \textbf{×} & 38.52 & 37.1 & 29.7 & 41.2 \\
     & \textbf{\checkmark} & \textbf{\checkmark} & \textbf{\checkmark} & \textbf{×} & 29.20 & 48.3 & 45.7 & 49.0 \\
     & \textbf{\checkmark} & \textbf{\checkmark} & \textbf{\checkmark} & \textbf{\checkmark} & 63.06 & \textbf{50.3} & \textbf{48.6} & \textbf{51.8} \\ \hline
    \end{tabular}
    }
    \caption{Effectiveness of the prompting mode and crucial components. "$\checkmark^{*}$" represents directly using support images for fine-tuning, \#Par(M) denotes the number of trainable parameters. Best results are marked in \textbf{bold}.}
    \vskip -0.08in
    \label{tab:ab-prompting}
\end{table}

\section{Conclusion}
In this paper, we introduce VisTex-OVLM, a novel object detection approach that uses visual textualization to project visual exemplars into the text feature space to prompt an OVLM without disrupting its pre-trained object-text alignment. To implement visual textualization, we design MSTBs and an MSF strategy that fully leverage OVLM's strong object-text alignment and object-level feature extraction capabilities, allowing the textualized visual tokens to achieve optimal semantic representation. VisTex-OVLM achieves state-of-the-art performance on both open-set detection (LVIS and 16 unseen datasets) and few-shot benchmarks (PASCAL VOC and MSCOCO). VisTex-OVLM is applicable to common OVLMs including GLIP and GroundingDINO, underscoring the benefits of using image prompts to complement text prompts and validating its effectiveness in addressing zero-shot object detection limitations. 
% Our work highlights visual textualization as a generalizable enhancement for OVLMs in real-world detection tasks.

% \vskip -0.28in
\paragraph{Acknowledgement} Supported by the National Natural Science Foundation in China under Grant 62371016 and U23B2063, the Bejing Natural Science Foundation Haidian District Joint Fund in China under Grant L222032, the Academic Excellence Foundation of BUAA for PhD Students.

\clearpage

{
    \small
    \bibliographystyle{ieeenat_fullname}
    \bibliography{main}
}

\clearpage
\setcounter{page}{1}
\maketitlesupplementary

% \section{Rationale}
% \label{sec:rationale}
% % 
% Having the supplementary compiled together with the main paper means that:
% % 
% \begin{itemize}
% \item The supplementary can back-reference sections of the main paper, for example, we can refer to \cref{sec:intro};
% \item The main paper can forward reference sub-sections within the supplementary explicitly (e.g. referring to a particular experiment); 
% \item When submitted to arXiv, the supplementary will already included at the end of the paper.
% \end{itemize}
% % 
% To split the supplementary pages from the main paper, you can use \href{https://support.apple.com/en-ca/guide/preview/prvw11793/mac#:~:text=Delete%20a%20page%20from%20a,or%20choose%20Edit%20%3E%20Delete).}{Preview (on macOS)}, \href{https://www.adobe.com/acrobat/how-to/delete-pages-from-pdf.html#:~:text=Choose%20%E2%80%9CTools%E2%80%9D%20%3E%20%E2%80%9COrganize,or%20pages%20from%20the%20file.}{Adobe Acrobat} (on all OSs), as well as \href{https://superuser.com/questions/517986/is-it-possible-to-delete-some-pages-of-a-pdf-document}{command line tools}.

\section{Implementation details of VisTex-DINO and other comparison methods}

We also implemented VisTex-OVLM on GroundingDINO-T \cite{liu2023groundingdino}, denoted as VisTex-DINO. The MSTB design mirrors that of VisTex-GLIP, employing two "fully connected (fc) + ReLU" layers. Based on the feature size and scale extracted by the pre-trained GroundingDINO-T's vision encoder, $H$ is set to 100. Since the intermediate features of GroundingDINO have $W$ values that vary with image size, we applied bilinear interpolation to also set $W$ to 100. GroundingDINO's intermediate features are available at three scales, making $M=3$. MSTB is applied to both the image backbone and the feature enhancer of GroundingDINO-T, with max pooling across stages used for non-parametric multi-stage fusion. All other training settings are consistent with those of VisTex-GLIP.

% 至于正文中comparison methods，We used recent top-performing approaches for comparison, adopting published results where available and reproducing results with recommended settings on PASCAL VOC and MSCOCO when needed. 需要说明的是，为了公平比较，其中OWL-ViT和OWL-ViT v2使用的权重版本分别是"CLIP ViT-L/14"和“CLIP B/16 ST+FT”，因为这两个权重在预训练过程中没有刻意剔除MSCOCO和LVIS中可能出现的类别，接近于我们使用的OVLM的权重的预训练设置。
\textcolor{black}{For the comparison methods, we selected recent top-performing approaches, adopting published results where available and reproducing results with recommended settings on PASCAL VOC and MSCOCO when necessary. Notably, to ensure a fair comparison, the weights used for OWL-ViT and OWL-ViT v2 are "CLIP ViT-L/14" and "CLIP B/16 ST+FT," respectively. These weights were chosen because they were not specifically trained to exclude categories that might appear in MSCOCO or LVIS, making their pre-training settings closer to those of the OVLM weights we used.} \textcolor{black}{We used class names as the text prompt for all methods that accept text prompts.}

\section{Performance on ODinW13 subsets}

Following \cref{sec:4.4} of the main text, we provide detailed transfer results on the ODinW13 subsets \cite{li2022glip} in \cref{tab:odinwresult}. ODinW13 \cite{li2022odinw} is composed of 13 subsets from ODinW35, spanning specialized natural domains such as aquarium species, surgical instruments, and aerial imagery, among others. Although the categories in these 13 datasets may appear in the pre-training dataset of OVLM, their performance results still, to some extent, reflect the model's capability in real-world scenarios. The specifics of ODinW13 are outlined in \cref{tab:odinwsetting}. All methods, including ours, were trained on the MSCOCO base set, treating downstream task sets as novel sets and evaluating them with 2-shot support images. The results are presented in the table below. These results further validate VisTex-OVLM's superior transferability across diverse domains, demonstrating its robustness and adaptability in handling significant domain shifts.

\section{Compatibility experiments on RegionCLIP and FIBER} We evaluated VisTex on RegionCLIP \cite{zhong2022regionclip} under a one-shot setting using the Open-Vocabulary COCO and LVIS benchmark, where base and novel categories are disjoint (\cref{regionclip}). The zero-shot (ZS) results were adopted from the original paper. In full fine-tuning (FF*), the model was fine-tuned with support images from both base and novel classes. In contrast, VisTex was trained only on base categories. Results show that VisTex still improves RegionCLIP’s performance on novel categories.  Furthermore, we tested VisTex on another object-level VLM, FIBER \cite{dou2022fiber}, under the same setting as Table 1 in the main text, and present the results in \cref{open-set} to further demonstrate its generalization ability.
% The smaller gain compared to GLIP is mainly due to RegionCLIP's lack of a multi-scale encoder (see Response 2).
\begin{table}[h]
% \vspace{-10pt}
\caption{Performance on RegionCLIP \cite{zhong2022regionclip}.}
% \vspace{-10pt}
\centering
\label{regionclip}
\resizebox{\columnwidth}{!}{
    \begin{tabular}{c|ccc|cc}
    \hline
    \multirow{2}{*}{\textbf{Method}} & \multicolumn{3}{c|}{\textbf{COCO}} & \multicolumn{2}{c}{\textbf{LVIS}} \\ \cline{2-6} 
     & \textbf{Novel AP50} & \textbf{Base AP50} & \textbf{All AP50} & \textbf{AP} & \textbf{APr} \\ \hline
    \textbf{regionCLIP-ZS} & 39.3 & 61.6 & 55.7 & 32.3 & 22 \\
    \textbf{regionCLIP-FF*} & 61.1 & 62.7 & 61.9 & 45.2 & 36.1 \\
    \textbf{VisTex-regionCLIP} & \textbf{65.3} & \textbf{68.2} & \textbf{67.4} & \textbf{47.8} & \textbf{40.3} \\ \hline
    \end{tabular}
}
% \vspace{-10pt}
\end{table}

\begin{table}[h]
\vspace{-10pt}
\caption{Performance on FIBER \cite{dou2022fiber} (mAP if not specified).}
\vspace{-10pt}
\centering
\label{open-set}
\resizebox{\columnwidth}{!}{
    \begin{tabular}{c|cc|ccccc}
    \hline
    \multirow{2}{*}{\textbf{Method}} & \multicolumn{2}{c|}{\textbf{LVIS MiniVal}} & \multicolumn{5}{c}{\textbf{Unseen   medical datasets}} \\ \cline{2-8} 
     & \textbf{AP} & \textbf{APr} & \textbf{MoNu} & \textbf{CCRCC} & \textbf{ConSeP} & \textbf{LIDC} & \textbf{Deeplesion} \\ \hline
    \textbf{FIBER-ZS} & 35.8 & 29.5 & 0.3 & 0.6 & 1.3 & 0.1 & 0.3 \\
    \textbf{FIBER-FF*} & 48.4 & 38.8 & 9.2 & 9.8 & 25.5 & 29.5 & 34.7 \\
    \textbf{VisTex-FIBER} & \textbf{49.6} & \textbf{41.6} & \textbf{10.5} & \textbf{11.4} & \textbf{26.8} & \textbf{32.9} & \textbf{36.4} \\ \hline
    \end{tabular}
}
\vspace{-10pt}
\end{table}

\section{Computational Overhead and Preprocess Time} In \cref{computational}, we report the computational overhead for processing one image using GLIP-L on RTX3090 with one support image, comparing it to MQ-Det and GLIP-FF. After an initial preprocessing step on the support image, textualized visual tokens are stored for reuse. Thus, the actual inference cost, aside from this initial preprocessing, remains identical to that of the original OVLM. For fair and direct comparison, the FLOPs and time corresponding to this one-time preprocessing step are highlighted in \textcolor{blue}{blue} in the table.

\begin{table}[t]
% \vspace{-6pt}
\caption{Computational Overhead (Preprocessing costs: \textcolor{blue}{blue}).}
% \vspace{-10pt}
\centering
\label{computational}
\resizebox{0.9\columnwidth}{!}{
    \begin{tabular}{c|cc|c|c}
    \hline
    \multirow{2}{*}{\textbf{Method}} & \multicolumn{2}{c|}{\textbf{FLOPs}} & \multirow{2}{*}{\textbf{\#Param(Trainable)}} & \multirow{2}{*}{\textbf{Inference time}} \\
     & \textbf{Training} & \textbf{Inference} &  &  \\ \hline
    \textbf{MQ-Det} & 717.46G & 243.25G & 10.87M & 0.553s \\
    \textbf{GLIP-FF*} & 653.15G & 218.78G & 397.59M & 0.547s \\
    \textbf{VisTex-GLIP} & 702.11G & \textcolor{blue}{17.75G}+218.78G & 8.38M & \textcolor{blue}{0.031s}+0.547s \\ \hline
    \end{tabular}
}
% \vspace{-10pt}
\end{table}

% Following \cref{sec:4.4} of the main text, we provide the detailed transfer results on ODinW13 subsets \cite{li2022glip} in \cref{tab:odinwresult}.  ODinW13 \cite{li2022odinw} 是ODinW35的13个子集构成， spans 13 specialized natural domains, such as aquarium species, surgical instruments, and aerial imagery, etc. 尽管这13个数据集中的类别可能出现在OVLM的预训练数据集中，在他们上面的表现结果仍然在某些程度上可以展现模型在real-world scenario下的表现。 The specifics of ODinW13 are outlined in \cref{tab:odinwsetting}. All methods, including ours, were trained on the MSCOCO base set, treating downstream task sets as novel sets and evaluating them with 2-shot support images. The results are presented in the table below. These results further validate VisTex-OVLM's superior transferability across diverse domains, demonstrating its robustness and adaptability in handling significant domain shifts.

\begin{table}[h]
    \centering
    \resizebox{\columnwidth}{!}{
    \renewcommand\arraystretch{1}
    \begin{tabular}{c|cccc}
    \hline
    \textbf{Dataset} & \textbf{Objects of interest} & \textbf{Train} & \textbf{Val} & \textbf{Test} \\ \hline
    PascalVOC & Common objects (PascalVOC 2012) & 13690 & 3422 & \textbackslash{} \\
    AerialDrone & Boats, cars, etc. from drone images & 52 & 15 & 7 \\
    Aquarium & Penguins, starfish, etc. in an aquarium & 448 & 127 & 63 \\
    Rabbits & Cottontail rabbits & 1980 & 19 & 10 \\
    EgoHands & Hands in ego-centric images & 3840 & 480 & 480 \\
    Mushrooms & Two kinds of mushrooms & 41 & 5 & 5 \\
    Packages & Delivery packages & 19 & 4 & 3 \\
    Raccoon & Raccoon & 150 & 29 & 17 \\
    Shellfish & Shrimp, lobster, and crab & 406 & 116 & 58 \\
    Vehicles & Car, bus, motorcycle, truck, and ambulance & 878 & 250 & 126 \\
    Pistols & Pistol & 2377 & 297 & 297 \\
    Pothole & Potholes on the road & 465 & 133 & 67 \\
    Thermal & Dogs and people in thermal images & 142 & 41 & 20 \\ \hline
    \end{tabular}
    }
    \caption{The objects of interest for each subset and the image number of each split in ODinW13.}
    \label{tab:odinwsetting}
\end{table}

\begin{table*}[t]
    \centering
    \resizebox{\textwidth}{!}{
    \renewcommand\arraystretch{1}
    \begin{tabular}{c|ccccccccccccc|c}
    \hline
    \textbf{Method} & \textbf{AerialDrone} & \textbf{Vehicles} & \textbf{Aquarium} & \textbf{Mushrooms} & \textbf{Raccoon} & \textbf{Packages} & \textbf{Pothole} & \textbf{Shellfish} & \textbf{Rabbits} & \textbf{Pistols} & \textbf{Egohands} & \textbf{Pascalvoc} & \textbf{Thermal} & \textbf{Mean} \\ \hline
    \textbf{Meta-DETR} & 16.5 & 32.5 & 18.2 & 36.1 & 34.2 & 31.4 & 22.8 & 28.2 & 36.6 & 34.1 & 35.2 & 30.1 & 37.1 & 30.2 \\
    \textbf{DiGeo} & 22.6 & 37.3 & 25.1 & 49.2 & 41.4 & 44.2 & 26.7 & 39.8 & 47.0 & 38.7 & 41.6 & 42.2 & 39.7 & 38.1 \\
    \textbf{DeFRCN} & 23.5 & 36.4 & 24.8 & 43.8 & 44.1 & 44.9 & 25.6 & 40.1 & 45.3 & 39.0 & 43.5 & 50.9 & 35.4 & 38.3 \\
    \textbf{MFD} & 24.5 & 38.7 & 29.4 & 46.2 & 45.1 & 50.1 & 28.4 & 38.4 & 48.4 & 43.5 & 46.3 & 50.5 & 42.2 & 40.9 \\
    \textbf{MQ-Det} & 21.2 & 36.5 & 30.0 & 45.3 & 41.1 & 41.6 & 28.3 & 37.5 & 46.9 & 44.1 & 39.7 & 48.4 & 40.3 & 38.5 \\
    \textbf{VisTex-GLIP} & \textbf{27.4} & \textbf{41.2} & \textbf{33.3} & \textbf{50.7} & \textbf{46.7} & \textbf{55.2} & \textbf{30.1} & \textbf{46.0} & \textbf{49.6} & \textbf{51.1} & \textbf{55.9} & \textbf{51.2} & \textbf{47.8} & \textbf{45.1} \\ \hline
    \end{tabular}
    }
    \caption{Detailed 2-shot transfer results on ODinW13 subsets. The best values are highlighted in \textbf{bold}.}
    \label{tab:odinwresult}
\end{table*}

\section{More ablation and visualization results}

\begin{table}[t]
    \centering
    \resizebox{0.9\linewidth}{!}{
    \begin{tabular}{c|cc|ccc}
    \hline
    \multirow{2}{*}{\textbf{MSTB}} & \multirow{2}{*}{\textbf{Scale ($j$)}} & \multirow{2}{*}{\textbf{{$\Delta$} \#Par(M)}} & \multicolumn{3}{c}{\textbf{2-shot}} \\ \cline{4-6} 
     &  &  & \textbf{AP} & \textbf{bAP} & \textbf{nAP} \\ \hline
    \multirow{5}{*}{\textbf{w/o sharing}} & \textbf{0} & -8.34 & 47.1 & 47.0 & 47.1 \\
     & \textbf{0+1} & -7.36 & 48.8 & \textbf{48.7} & 49.2 \\
     & \textbf{0+1+2} & -3.68 & 48.6 & 47.9 & 48.8 \\
     & \textbf{0+1+2+3} & 1.23 & 48.9 & 47.0 & 49.5 \\
     & \textbf{0+1+2+3+4} & 5.15 & 49.2 & 48.5 & 50.6 \\ \hline
    \textbf{sharing} & \textbf{0+1+2+3+4} & 0.00 & \textbf{50.3} & 48.6 & \textbf{51.8} \\ \hline
    \end{tabular}
    }
    \caption{Ablation on multi-scale textualization and MSTB sharing. {$\Delta$}\#Par(M) represents the parameter amount offset relative to the optimal configuration. Best results are marked in \textbf{bold}.}
    \vskip -0.08in
    \label{tab:MSTB}
\end{table}

\begin{table}[t]
    \centering
    \resizebox{0.6\linewidth}{!}{
    \begin{tabular}{c|c|ccc}
    \hline
    \multirow{2}{*}{\textbf{Stages}} & \multirow{2}{*}{\textbf{$\Delta$\#Par(M)}} & \multicolumn{3}{c}{\textbf{2-shot}} \\ \cline{3-5} 
     &  & \textbf{AP} & \textbf{bAP} & \textbf{nAP} \\ \hline
    \textbf{1} & \textbf{-30.67} & 48.5 & 47.1 & 50.7 \\
    \textbf{1→4} & \textbf{-16.93} & 48.8 & 47.7 & 49.3 \\
    \textbf{1→8} & \textbf{0.00} & \textbf{50.3} & \textbf{48.6} & \textbf{51.8} \\
    \textbf{5→8} & \textbf{-16.93} & 23.2 & 30.7 & 13.6 \\
    \textbf{8} & \textbf{-30.67} & 20.5 & 28.9 & 13.9 \\ \hline
    \end{tabular}
    }
    \caption{Ablation on mapping and fusing different numbers and sequences of stages. "i→j" indicates the stages used. {$\Delta$}\#Par(M) represents the parameter amount offset relative to the optimal configuration. Best results are marked in \textbf{bold}.}
    \label{tab:stagefus}
    \vskip -0.15in
\end{table}

\subsection{Multi-scale textualizing block}
We assessed the impact of multi-scale textualization and the parameter-sharing strategy (MSTB sharing), as shown in \cref{tab:MSTB}. With scales indexed by $j$ (e.g., "0" represents the $ \frac{H}{2^0} \cdot \frac{W}{2^0} $ scale), results indicate that using multi-scale features from the vision encoder better preserves OVLM's object-text alignment and boosts performance over single-layer features. MSTB sharing further reduces the required convolutional weights, enhancing textualization effectiveness. MSTB sharing creates synergy during training, easing the learning process for mapping features of various scales into the same text feature space and slightly improving performance,. MSTB sharing saves 5.15M parameters and improves nAP by 1.2\% compared to non-sharing across scales.

%When MSTB sharing is employed, it reduces the convolutional weights that need to be stored and further improves the effectiveness of textualization. Using the same set of structures for textualizing mapping across different scale features exerts a synergistic effect during training, reducing the learning difficulty of the mapping function that maps features of different scales into the same text feature space, resulting in a slight performance improvement. Compared to using all scales without weight sharing, our final MSTB sharing saves 5.15M parameters and achieves a 1.2\% improvement in nAP.
\subsection{Multi-stage fusion}
Multi-stage fusion (MSF) merges features from multiple encoder stages into a single textualized visual token. The original GLIP has 8 stages. To maintain GLIP's object-text alignment, we map each stage’s visual features into the BERT-derived text feature space using MSTB. As shown in \cref{tab:stagefus}, using stages $1\rightarrow 8$ yields the best performance, while stages 1 and $1\rightarrow 4$ perform slightly lower. However, stages $5\rightarrow 8$ or just stage 8 result in significant drops, likely due to that as the neural network progresses, the features become more high-level and abstract, making the mapping learning more challenging. Continuously incorporating information from different stages starting from the lower layers helps reduce the difficulty of learning the mapping. 

\subsection{Ablation on shot fusion mode}
\textcolor{black}{When integrating multiple shot features, there are various fusion modes available. \cref{tab:shotfus} presents the performance of different fusion approaches. Experimental results indicate that concatenation yields the best results, as it maximally preserves the information from all shots, thereby preventing information loss. Support samples provide detailed semantic guidance for query prediction, and concatenation allows the model to maintain the unique information of each shot while utilizing data from multiple support samples. This method helps GLIP better grasp the support sample distribution for query prediction, enhancing performance.}

\begin{table}[t]
    \centering
    \resizebox{0.7\linewidth}{!}{
    \begin{tabular}{c|ccc}
    \hline
    \multirow{2}{*}{\textbf{Shot Fusion Mode}} & \multicolumn{3}{c}{\textbf{2-shot}} \\ \cline{2-4} 
     & \textbf{AP} & \textbf{bAP} & \textbf{nAP} \\ \hline
    \textbf{element-wise addition} & 41.6 & 41 & 41.9 \\
    \textbf{max} & 48.5 & 48.4 & 48.6 \\
    \textbf{average} & 47.9 & 45.2 & 48.1 \\
    \textbf{concat} & \textbf{50.3} & \textbf{48.6} & \textbf{51.8} \\ \hline
    \end{tabular}
    }
    \caption{Ablation on shot fusion mode. Best results are marked in \textbf{bold}.}
    % \vskip -0.08in
    \label{tab:shotfus}
\end{table}

\subsection{Ablation on MSF}
\textcolor{black}{MSF’s innovation is its efficient use of OVLM’s multi-stage object-text alignment without extra parameters. \cref{tab:stagefus} in the main text shows multi-stage fusion’s effectiveness. We conducted an MSF ablation study in \cref{MSF} using different common fusion methods. Max pooling outperforms other non-parametric fusion methods. This is because max pooling can highlight the most informative features across stages while reducing the negative impact of redundant noise.}

\begin{table}[h]
% \vspace{-10pt}
% \vspace{-10pt}
\centering
\label{MSF}
\resizebox{0.7\columnwidth}{!}{
    \begin{tabular}{c|ccc}
    \hline
    \multirow{2}{*}{\textbf{Stage Fusion Mode}} & \multicolumn{3}{c}{\textbf{2-shot}} \\ \cline{2-4} 
     & \textbf{AP} & \textbf{bAP} & \textbf{nAP} \\ \hline
    \textbf{element-wise Addition} & 36.7 & 37.4 & 31.6 \\
    \textbf{max} & \textbf{50.3} & \textbf{48.6} & \textbf{51.8} \\
    \textbf{average} & 46.8 & 48.3 & 41.4 \\
    \textbf{concat} & 39.4 & 39.3 & 39.4 \\ \hline
    \end{tabular}
}
\caption{Ablation on MSF. Best results are marked in \textbf{bold}.}
% \vspace{-13pt}
\end{table}

\subsection{Ablation on image prompt engineering}
\begin{table}[t]
    \centering
    \resizebox{0.7\linewidth}{!}{
    \begin{tabular}{c|c}
    \hline
    \multirow{2}{*}{\textbf{Engineering Method}} & \textbf{2-shot} \\ \cline{2-2} 
     & \textbf{AP} \\ \hline
    \textbf{baseline} & 16.3 \\
    \textbf{BG blur} & \textbf{49.6} \\
    \textbf{dye object red in grays image} & 19.5 \\
    \textbf{add red object outline} & 28.3 \\
    \textbf{crop} & 44.2 \\
    \textbf{crop large context} & 30.8 \\ \hline
    \end{tabular}
    }
    \caption{Ablation on image prompt engineering methods. “Baseline” indicates directly inputting the original image. "crop" means cropping out the target region based on ground truth bounding box while "crop large context" enlarges ground truth bounding box by $k=10$ pixels. "BG blur" technique applies a shadow (intensity of 0.1) and Gaussian noise (kernel size of 15 and a standard deviation of 3) to the background area. Best results are marked in \textbf{bold}.}
    \label{tab:Iengineer}
\end{table}

This ablation study (\cref{tab:Iengineer}) followed the same experimental setup outlined in the main text: (1) Conducted on VisTex-OVLM in a 2-shot setting on MSCOCO using GLIP-L; (2) mAP was reported for both base and novel classes, with all other settings kept optimal. Several methods for preprocessing and inputting image prompts were tested, following CLIPSeg \cite{luddecke2022clipseg} settings unless specified. Experimental results show that the "BG blur" technique performs best. It highlights the target object while preserving some background, unlike "crop" and "baseline." Additionally, it avoids overlaying original image pixels, preventing information loss.

\subsection{More output visualization}

\begin{figure}[h]
    \centering
    \includegraphics[width=\linewidth]{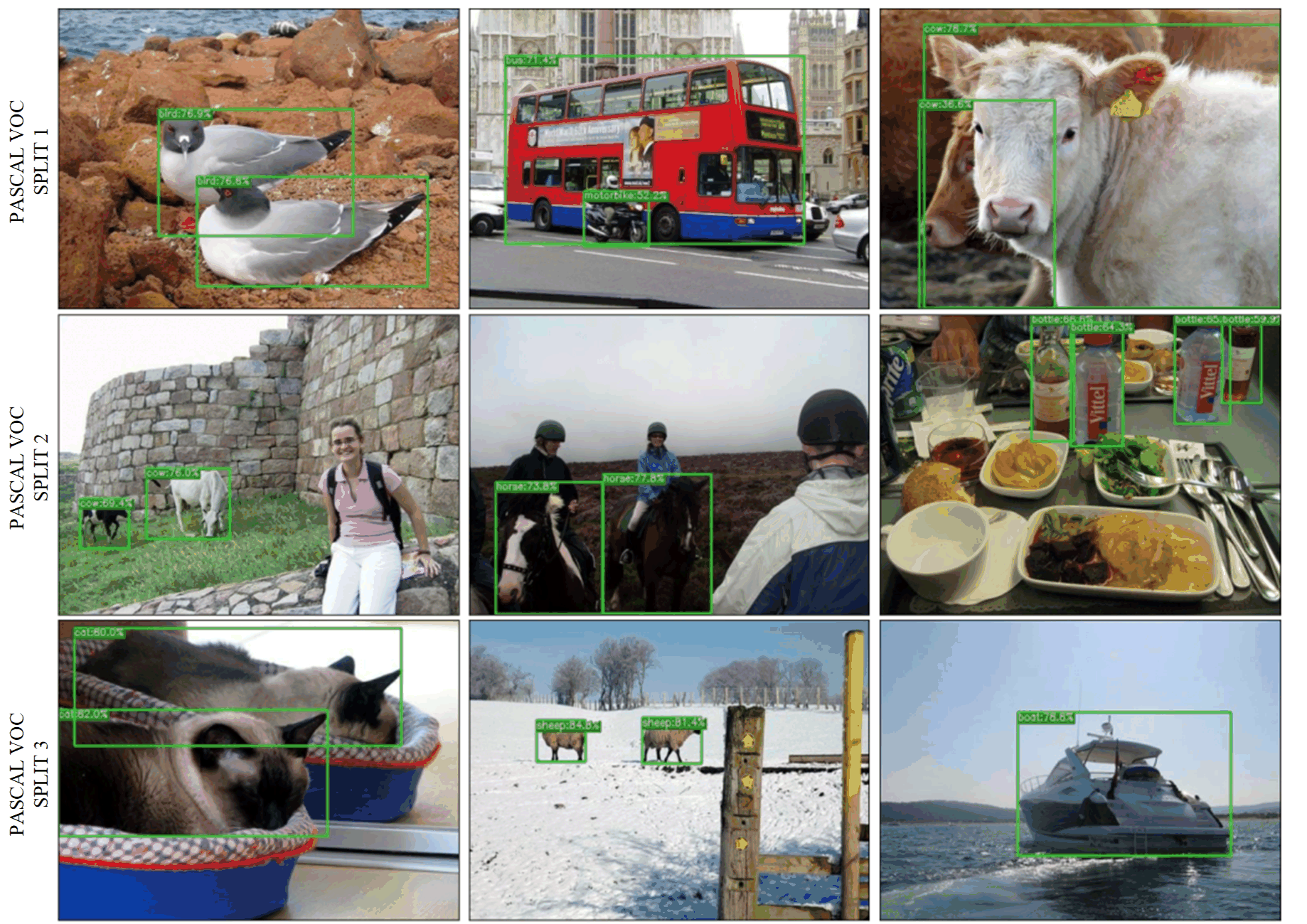}
    \caption{Visualization of VisTex-GLIP’s 10-shot object detection results on PASCAL VOC. For simplicity, only detections of novel-class objects are illustrated. The settings are corresponded to \cref{tab:pascal-com} in the main text.}
    \label{fig:pascal}
\end{figure}

We provide 10-shot output examples for VisTex-GLIP on PASCAL VOC in \cref{fig:pascal}. The settings are corresponded to \cref{tab:pascal-com} in the main text.

\begin{figure}[t]
    \centering
    \includegraphics[width=\linewidth]{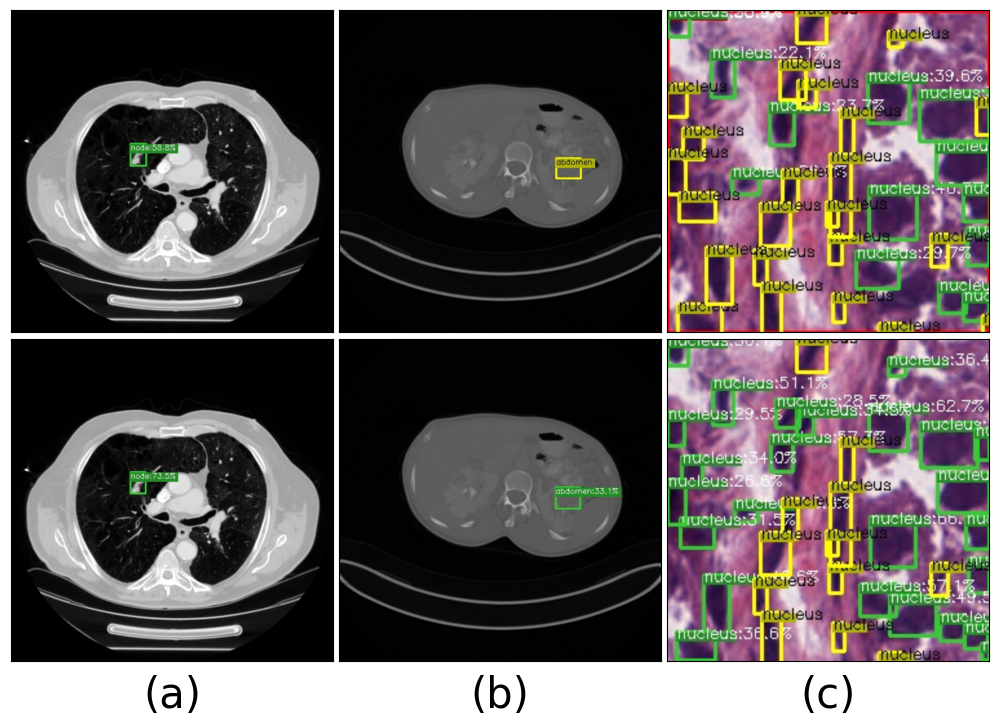}
    \caption{Output visualizations for medical datasets. The first row illustrates the inference results when models trained solely on the MSCOCO base set were directly applied to the medical datasets. The second row demonstrates results after briefly fine-tuning the MSTB in a 2-shot setting on the medical tasks. (a) LIDC, (b) Deepleision, (c) MoNu. Green, red, and yellow boxes denote true positives, false positives, and false negatives. }
    \label{fig:med}
\end{figure}

\begin{figure}[t]
    \centering
    \includegraphics[width=\linewidth]{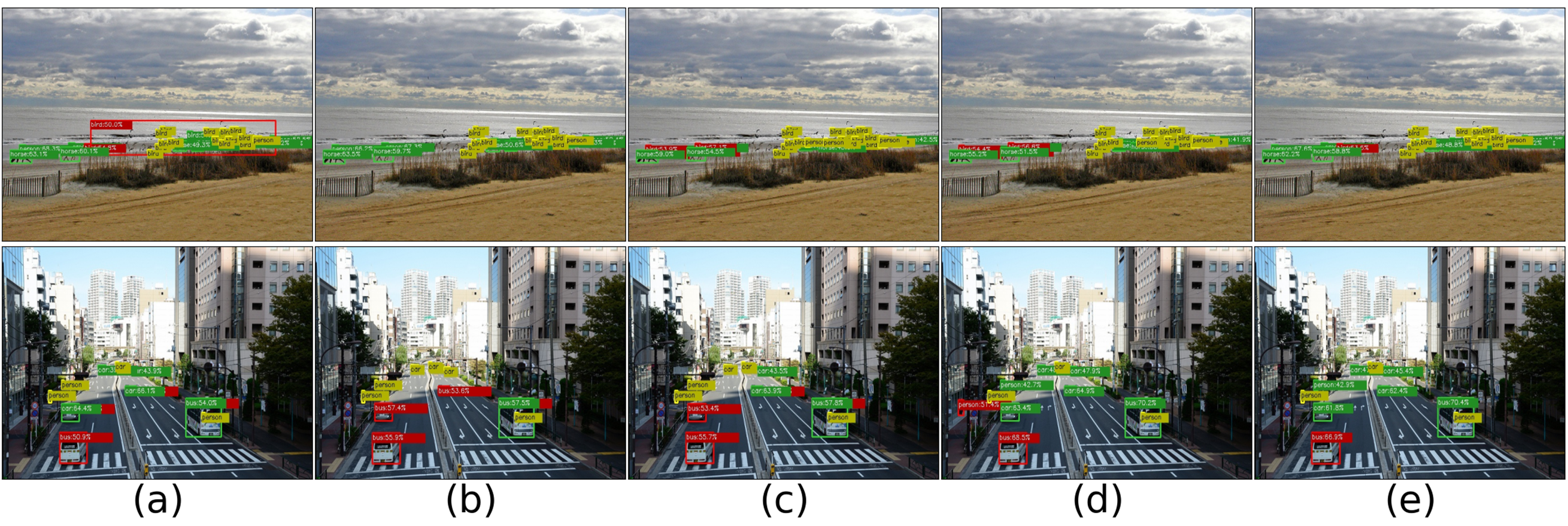}
    \caption{Failure cases. (a) GLIP-ZS, (b) GLIP-FF, (d) GLIP-MaPLe, (e) MQ-Det, (f) VisTex-GLIP. Green, red, and yellow: true positives, false positives, and false negatives.}
    \label{fig:failure}
\end{figure}

\section{Output visualization for real-world downstream tasks}

\begin{figure*}
    \centering
    \includegraphics[width=0.61\textwidth]{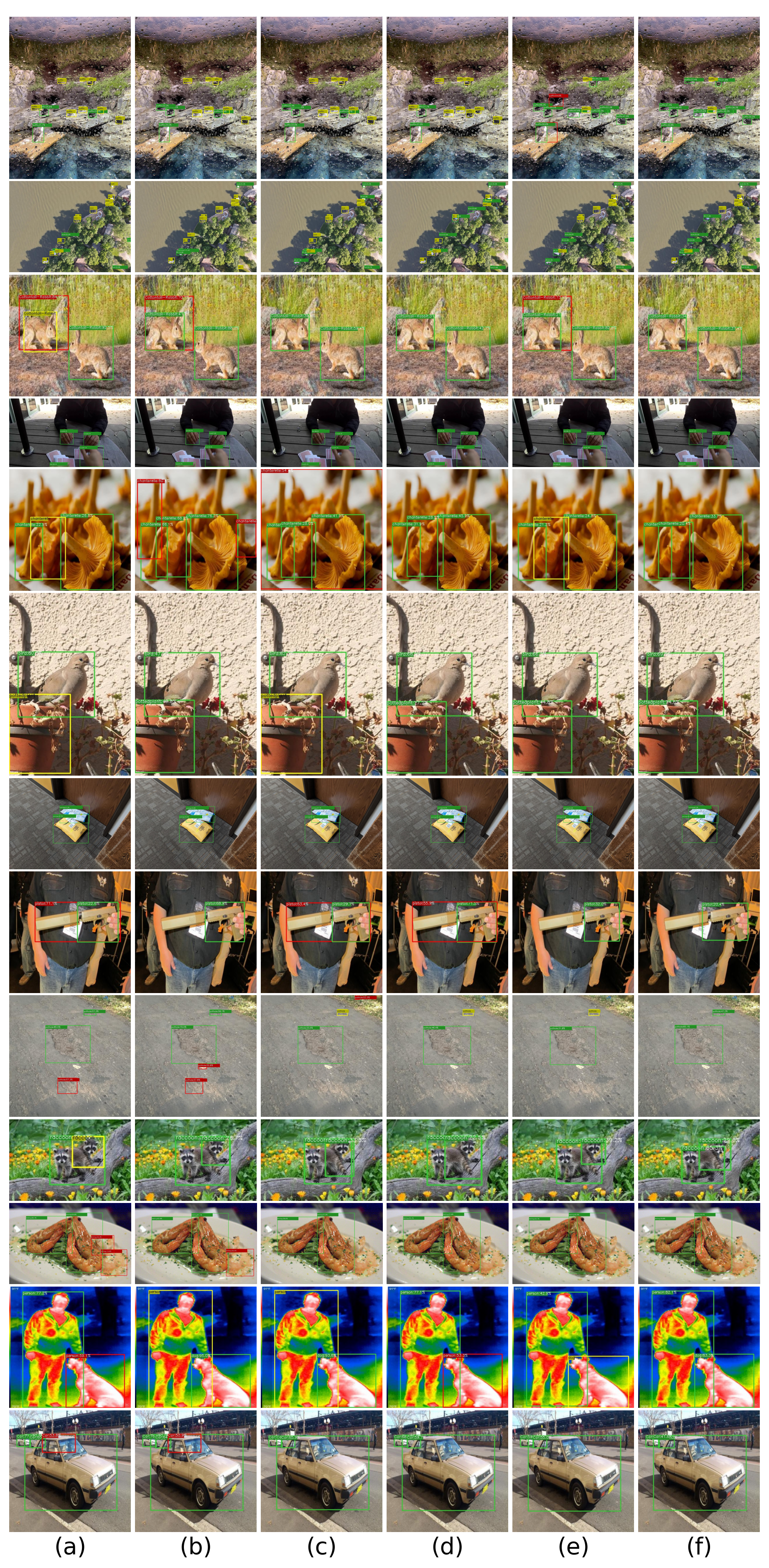}
    \caption{Comparison output visualizations on ODinW13. From top row to the bottom: Aquarium, AerialDrone, Rabbits, EgoHands, Mushrooms, PascalVOC, Packages, Pistols, Pothole, Raccoon, Shellfish, Thermal, Vehicles. (a) Meta-DETR, (b) DiGeo, (c) DeFRCN, (d) MFD, (e) MQ-Det, (f) VisTex-GLIP. Green, red, and yellow boxes denote true positives, false positives, and false negatives. }
    \label{fig:odinw}
\end{figure*}

\cref{fig:odinw} and \cref{fig:med} present the output visualizations for real-world downstream tasks, including ODinW13 subsets and medical datasets (MoNu, LIDC, and Deeplesion).

The visualizations in \cref{fig:odinw} adhere to the settings described in \cref{sec:4.4} of the main text, where all methods, including ours, were trained on the MSCOCO base set. Downstream task sets were treated as novel sets, and evaluations were conducted using 2-shot support images, directly showcasing inference results on novel classes.

\cref{fig:med} focuses on datasets with larger domain gaps, specifically medical datasets. The first row illustrates the inference results when models trained solely on the MSCOCO base set were directly applied to the medical datasets. The second row demonstrates results after briefly fine-tuning the MSTB in a 2-shot setting on the medical tasks.

\textcolor{black}{We also provide some failure cases of our method on natural images in \cref{fig:failure}. As shown, these failures primarily occur in scenarios with dense or small objects, which is similar to the challenges observed in medical images in \cref{fig:med}. We attribute this limitation to the weak representation of small objects in the pre-trained OVLM and the inherent difficulty of fitting novel category distributions with limited support samples. Addressing these issues will be a focus of future research.}

% 我们还在图18中给出了一些我们的方法在自然图像的failure cases。可以看到，这些failure primarily occur in scenarios with dense or small objects. 这和 \cref{fig:med}中展示的医学图像的状况类似。我们认为这个limitation源于pre-trained OVLM细小物体表征姣弱，以及少量support samples本身难以拟合novel categories 的distribution。如何解决这些问题将是未来的研究方向。

\end{document}